\RequirePackage{fix-cm}
\documentclass[twocolumn]{svjour3}          
\smartqed  
\usepackage{natbib}
\usepackage{times}
\usepackage{epsfig}
\usepackage{graphicx}
\usepackage{subfigure}
\usepackage{amsmath,amssymb,amsfonts,dsfont,pifont,bm,bbm,mathrsfs,mathtools,nicefrac}
\usepackage{bm}
\usepackage{textcomp}
\usepackage{multirow}
\usepackage{xspace}
\usepackage{xcolor}
\usepackage{booktabs}
\usepackage{tabularx}
\usepackage{color}
\usepackage{makecell}
\usepackage{wrapfig}
\usepackage{bbding}

\usepackage{algorithm}
\usepackage{algorithmic}

\definecolor{citecolor}{RGB}{119,185,0} 
\usepackage[pagebackref=false,breaklinks=true,colorlinks,citecolor=citecolor,bookmarks=false]{hyperref}

\newcommand{\revise}[1]{\textcolor{black}{#1}}
\definecolor{upcolor}{RGB}{57,182,74}
\newcommand{\up}[1]{\textcolor{upcolor}{$\uparrow$ #1}}
\newcommand{\down}[1]{\textcolor{red}{$\downarrow$ #1}}
\newlength\savewidth
\definecolor{Light}{RGB}{246,234,227}

\newcommand{\method}{T2S-DA\xspace}

\renewcommand{\paragraph}[1]{\vspace{1.25mm}\noindent\textbf{#1}}

\makeatletter
\DeclareRobustCommand\onedot{\futurelet\@let@token\@onedot}
\def\@onedot{\ifx\@let@token.\else.\null\fi\xspace}

\definecolor{deemph}{gray}{0.6}
\newcommand{\gc}[1]{\textcolor{deemph}{#1}}

\usepackage{soul}
\soulregister\cite7
\soulregister\eqref7
\soulregister\ref7

\begin{document}

\title{Pulling Target to Source: A New Perspective on Domain Adaptive Semantic Segmentation
}


\author{Haochen Wang$^{1,2}$ \and Yujun Shen$^4$ \and Jingjing Fei$^5$ \and Wei Li$^5$ \and Liwei Wu$^5$ \\ Yuxi Wang$^{3*}$ \and Zhaoxiang Zhang$^{1,2,3*}$
}


\institute{
    $^1$ New Laboratory of Pattern Recognition, State Key Laboratory of Multimodal Artificial Intelligence Systems, Institute of Automation, Chinese Academy of Sciences, Beijing, China.
    \\
    $^2$ University of Chinese Academy of Sciences, Beijing, China. \\
    $^3$ Centre for Artificial Intelligence and Robotics, Hong Kong Institute of Science \& Innovation, Chinese Academy of Sciences, Hong Kong, China. \\  
    $^4$ Chinese University of Hong Kong, Hong Kong, China. \\
    $^5$ SenseTime Research, Beijing, China. \\
    $^*$ Yuxi Wang and Zhaoxiang Zhang are the corresponding authors. \\
    E-mail: \{wanghaochen2022, zhaoxiang.zhang\}@ia.ac.cn, and yuxiwang93@gmail.com.
}

\date{Received: date / Accepted: date}

\maketitle

\begin{abstract}
Domain-adaptive semantic segmentation aims to transfer knowledge from a labeled source domain to an unlabeled target domain.
However, existing methods primarily focus on directly learning \revise{categorically discriminative} target features \revise{for segmenting target images, which is challenging} in the absence of target labels.
This work provides a new perspective.
We observe that the features learned with source data manage to keep categorically discriminative during training, thereby enabling us to implicitly learn adequate target representations by simply \textit{pulling target features close to source features for each category}.
To this end, we propose \method, which encourages the model to learn similar cross-domain features.
%
%
Also, considering the pixel categories are heavily imbalanced for segmentation datasets, we come up with a dynamic re-weighting strategy to help the model concentrate on those underperforming classes.
Extensive experiments confirm that \method learns a more discriminative and generalizable representation, significantly surpassing the state-of-the-art.
We further show that \method is quite qualified for the domain generalization task, verifying its domain-invariant property.
\keywords{Domain Adaptation \and Semantic Segmentation}
\end{abstract}

\section{Introduction}\label{sec:intro}

\begin{figure}[t]
    \centering
    \includegraphics[width=1.0\linewidth]{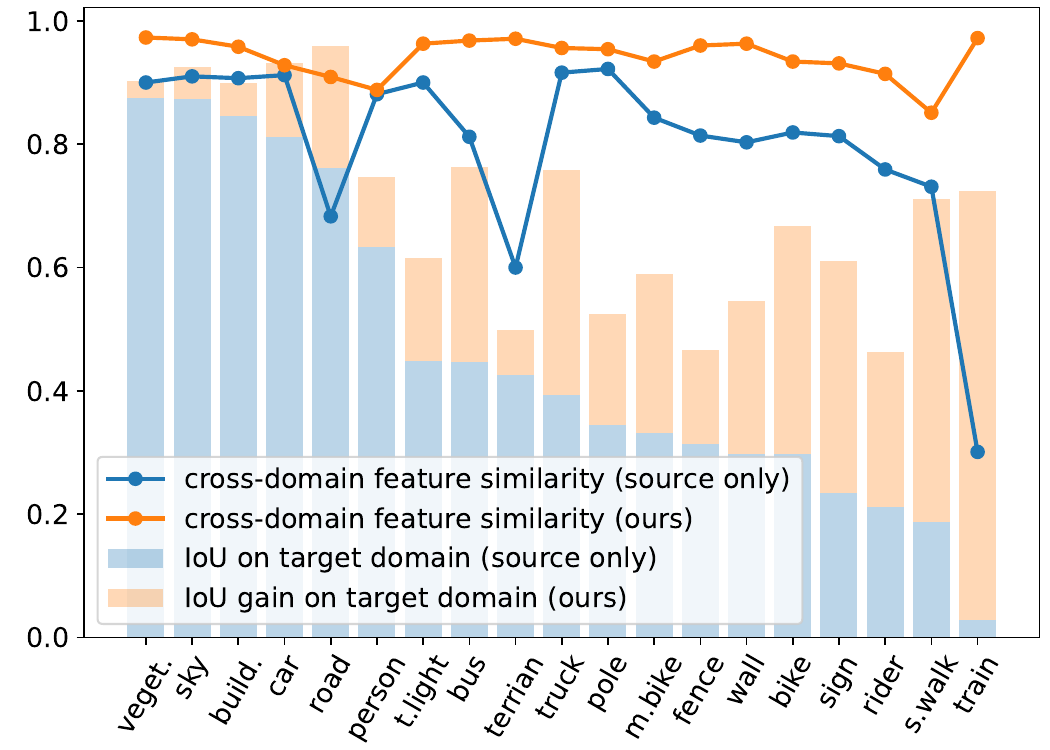}
    \caption{%
    \textbf{Category-wise cross-domain feature similarity as well as the evaluation results on the target domain}.
    %
    %
    When directly testing the model trained with source data (\textit{i.e.}, ``source only'') on the target data, the categories, where source and target features are \textit{largely dissimilar} to each other, suffer from low IoU.
    %
    %
    }
    \label{fig:stats}
\end{figure}

The success of semantic segmentation largely relies on big data \citep{long2015fully, ronneberger2015u, zhao2017pyramid}, however, collecting a sufficient number of annotated images could be labor-intensive in practice \citep{cordts2016cityscapes, zhou2017scene}.
Recent studies \citep{zhang2019category, kang2020pixel, araslanov2021self, hoyer2022daformer} yield an alternative solution by introducing synthetic datasets \citep{richter2016gta, ros2016synthiaa}, where the labels can be obtained with minor effort.
However, the models learned on these datasets are found hard to generalize to real-world scenarios.
To address this challenge, unsupervised domain adaptation (UDA) \citep{zou2018unsupervised, vu2019advent, lian2019constructing, chen2022reusing} is proposed to transfer knowledge from the labeled source domain (\textit{e.g.}, synthetic) to an unlabeled target domain (\textit{e.g.}, real).
Under this setting, the crux becomes how to make full use of source labels to learn discriminative representations for segmenting target samples.

To address this issue, typical solutions fall into two categories, \textit{i.e.}, adversarial training, and self-training.
The former tries to align the feature distribution of different domains to an \textit{intermediate common space} \citep{goodfellow2014generative, nowozin2016f, tsai2018learning, luo2019taking, chen2019learning, tsai2019domain,  pan2020unsupervised, wang2020differential, luo2021category, ganin2016domain, long2018conditional}.
Although these approaches converge the two domains globally, it is hard to ensure that target features from different categories are well-separated \citep{xie2022sepico, du2022learning}.
The latter aims to build a capable \textit{target} feature space by selecting confident target predictions as pseudo-ground truths \citep{tranheden2021dacs, zou2019confidence, zhang2019category, li2022class, zheng2021rectifying, araslanov2021self, shin2020two, du2022learning, yang2020fda, wang2023balancing, chen2022deliberated}.
These approaches \textit{directly} supervise the model with target pseudo-labels.
However, the segmentation model usually tends to be biased to the source domain, resulting in error-prone target pseudo-labels \citep{li2022class}.

We first explore the underlying causes of the performance drop \revise{and the consequences of the distribution shift} when applying the model trained on source data to target data.
As blue bars in Fig. \ref{fig:stats} suggest, \textit{not all} categories suffer in this domain shift.
Specifically, the performances of ``vegetation'', ``sky'', and ``building'' manage to remain satisfactory under the domain gap.
However, the performance of ``train'' deteriorates severely under this setting.
The main differences between these two types of categories are \textit{feature dissimilarities}.
\revise{
Concretely, features of those underperforming classes are \textit{dissimilar} across domains, \textit{i.e.}, with \textit{small} similarity.
}

\begin{figure}[t]
    \centering
    \includegraphics[width=1\linewidth]{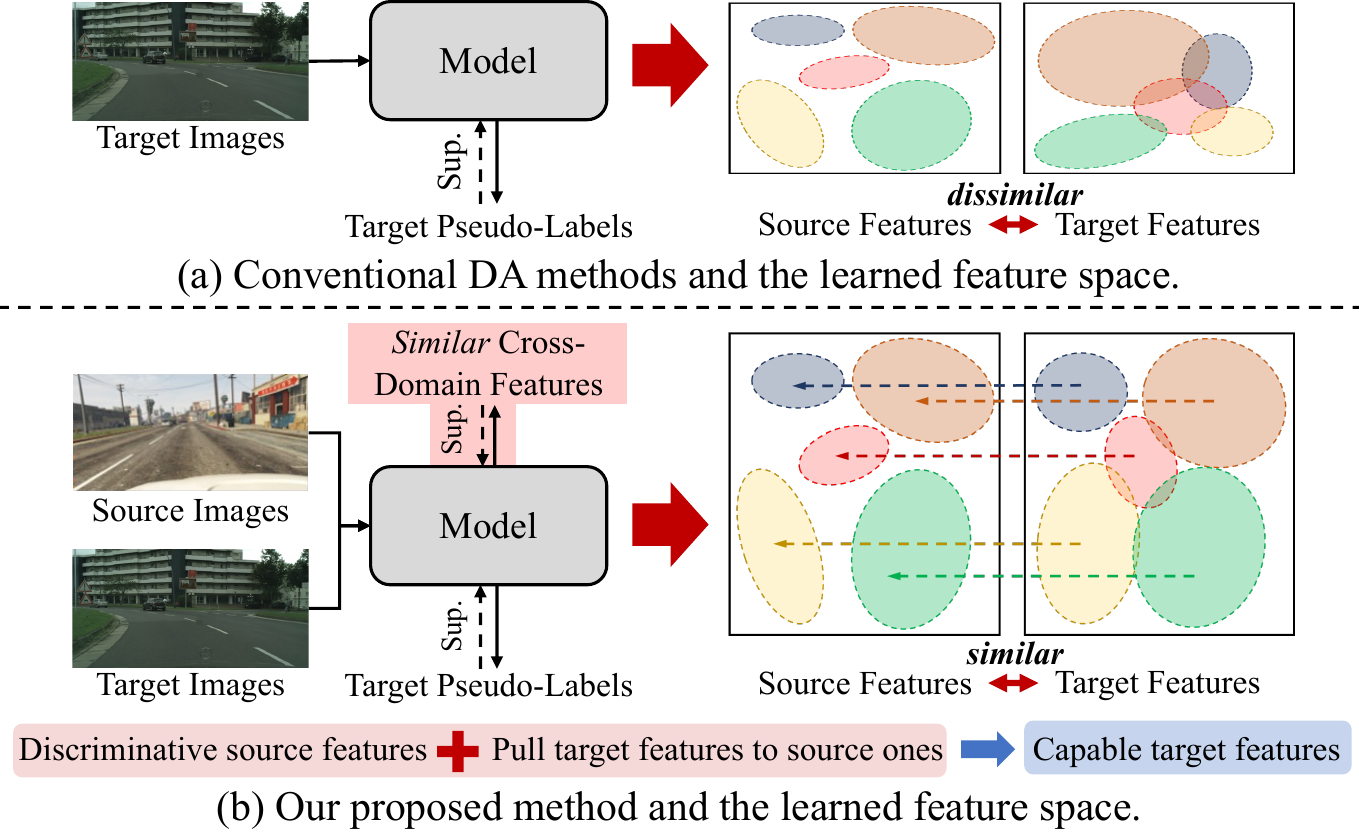}
    \caption{%
    \textbf{Concept comparison} between (a) conventional methods and (b) our \method.
    To obtain discriminative features from target images, existing approaches \textit{directly} supervise the model with target pseudo-labels regardless of the similarity between cross-domain features.
    Differently, \method addresses this issue from a new perspective, where we argue that ``discriminative source features'' plus ``making target features close to source features'' \textit{implicitly} brings capable target features.
    }
    \label{fig:motivation}
    \vspace{-10pt}
\end{figure}

Based on this evidence, we aim to design a framework that can extract \textit{similar} cross-domain features regarding each category.
Intuitively, source features are \textit{always} \revise{categorically discriminative} during training thanks to sufficient source labels. 
Therefore, as illustrated in Fig. \ref{fig:motivation}, we argue that (1) discriminative source features plus (2) urging target features to be similar to the source feature for each category, implicitly brings the categorical discriminativeness of target features.
To this end, we propose \method, regarding source features as \textit{anchors} and explicitly pulling target features \textit{close} to source ones.
However, due to the lack of target labels, it is hard to conduct feature pairs that \textit{exactly} belong to the same class.

To address this issue, we employ an image translation engine to produce pseudo-images by transferring source data to the target style.
These pseudo-images are then served as queries and can be easily matched with their positive keys (\textit{i.e.}, source features from the same category) \textit{precisely} since they naturally have annotations.
Additionally, considering the pixel categories are heavily imbalanced for segmentation datasets, we put forward a dynamic re-weighting strategy, forcing the model to put more effort into those underperforming classes.
Through this way, our approach is able to learn similar representations across domains and hence achieves substantial improvements on the target domain. 
From Fig. \ref{fig:stats}, we can tell that \textit{the improved similarity indeed contributes to better performances}, especially for class ``train''.

We evaluate the proposed \method on two UDA benchmarks, \textit{i.e.}, GTA5 \citep{richter2016gta} $\to$ Cityscapes \citep{cordts2016cityscapes} and Synthia \citep{ros2016synthiaa} $\to$ Cityscapes \citep{cordts2016cityscapes}, where we consistently surpass state-of-the-art alternatives.
For instance, \method achieves $75.1\%$ mIoU on GTA5 $\to$ Cityscapes benchmark, outperforming HRDA \citep{hoyer2022hrda} by $+1.3\%$.
Moreover, we find \method is also applicable to the domain generalization task, where the training phase cannot access the target samples at all.
Under this setting, when training the model on GTA5 and Synthia and testing on City-scapes, we improve the baseline by $+2.5\%$ and $+2.1\%$, respectively.


\section{Related Work}\label{sec:related}

\subsection{Domain Adaptive Semantic Segmentation}

Domain adaptive semantic segmentation aims at learning a generalized segmentation model that can adapt from a labeled (synthetic) source domain to an unlabeled (real-world) target domain. 
To overcome the domain gap, most previous methods align distributions of source and target domains to an \textit{intermediate common space} at the image level \citep{hoffman2018cycada, murez2018image, sankaranarayanan2018learning, li2019bidirectional},
feature level \citep{hoffman2016fcns, hong2018conditional, saito2018maximum, chang2019all, chen2019progressive}, 
and output level \citep{tsai2018learning, luo2019taking} by introducing extra objectives or techniques, 
\textit{e.g.}, optimizing some custom distance \citep{long2015learning, lee2019sliced}, 
applying computationally adversarial training \citep{goodfellow2014generative, nowozin2016f, tsai2018learning, luo2019taking, chen2019learning, tsai2019domain}, 
offline pseudo-labeling \citep{zou2018unsupervised, li2019bidirectional, zou2019confidence}, 
and image translation models \citep{hoffman2018cycada, li2019bidirectional}.
Different from these methods, as illustrated in Fig. \ref{fig:motivation}\textcolor{red}{b}, based on the observation that source features are always capable during training, we \textit{pull target features close to source features for each category}.
Through this way, \method manages to learn similar cross-domain features for each category.
Shown in Fig. \ref{fig:stats}, the improved cross-domain feature similarity indeed boosts the segmentation results.
One may be concerned that making the source close to the target seems to be a reasonable alternative.
We compare this strategy, \textit{i.e.}, ``source $\to$ target'', to our \method in Sec. \ref{sec:abla}, where we empirically find that this explicit alignment works \textit{only when pulling target to source}.

\subsection{Domain Generalized Semantic Segmentation}

Domain generalized semantic segmentation is a more challenging task compared to domain adaptation.
It assumes target data is unaccessible during training, focusing on generalizing well on \textit{unseen} target domains.
To extract domain-invariant feature representations, plenty of approaches have been proposed such as meta-learning \citep{li2018learning, balaji2018metareg, li2019episodic, li2019feature}, adversarial training \citep{li2018domain, li2018deep, rahman2020correlation}, metric-learning \citep{motiian2017unified, dou2019domain}, and feature normalization \citep{pan2018two, huang2019iterative, choi2021robustnet}.
Few attempts have been made in domain generalization based on cross-domain alignment.
\method is proved to be efficient for both settings, verifying that similar cross-domain features do contribute to better segmentation results.

\subsection{Contrastive Learning in Semantic Segmentation}
Recently, contrastive learning has been verified to be a successful framework for unsupervised or self-supervised representation learning in computer vision \citep{he2020momentum, grill2020bootstrap, li2021supervision, chen2020simple, chen2021exploring, caron2021emerging, wang2023bootstrap, wang2023droppos, wang2023hard, wang2024ross},
and has been explored in fully-supervised \citep{wang2021exploring, hu2021region}, semi-supervised \citep{wang2022semi, liu2022bootstrapping, wang2023using}, and weakly-supervised \citep{du2021weakly} semantic segmentation.
Only a few studies introduce contrastive learning in domain adaptive semantic segmentation.
Let us consider leveraging contrastive learning to minimize the domain gap.
It becomes crucial that positive pairs should be (1) matched precisely at the category level and (2) across domains, regardless of the level where the contrast is conducted.
In the literature, contrast is conducted at pixel \citep{kang2020pixel}, category \citep{huang2022category}, region \citep{zhou2021domain}, and history \citep{huang2021model} levels, respectively.
\method is the only one that both satisfies (1) and (2).
Specifically, it conducts contrast at the category level \textit{across domains}.
By employing an image translation engine, \method is a more efficient framework since it \textit{ensures cross-domain features are matched correctly} when conducting positive pairs.
We summarize these two properties in Tab.~\ref{tab:related}.
Interestingly, consistent improvements are observed by simply minimizing MSE between $\ell_2$ normalized positive pairs, verifying that contrastive learning is \textit{not} the only way of pulling target to source.

\begin{table}[t]
    \centering\footnotesize
    \setlength{\tabcolsep}{3pt}
    \caption{
    Comparison among domain adaptive semantic segmentation methods using contrastive learning.
    Our \method makes positive pairs are (1) matched precisely and (2) across domains.
    }
    \label{tab:related}
    \vspace{-5pt}
    \begin{tabular}{l|cc|cc}
    \toprule
    \multirow{3}{*}{Method} & \multirow{3}{*}{Queries} & \multirow{3}{*}{Positive Keys} & \multicolumn{2}{c}{Positive pairs are} \\
    \cline{4-5}
    & & & matched & across \\
    & & & precisely & domains \\
    \midrule
    \citep{kang2020pixel} & source & target & {\scalebox{0.7}{\XSolidBrush}} & \checkmark \\
    \citep{huang2022category} & target & source + target & {\scalebox{0.7}{\XSolidBrush}} & {\scalebox{0.7}{\XSolidBrush}} \\
    \citep{zhou2021domain} & target & target & \checkmark & {\scalebox{0.7}{\XSolidBrush}} \\
    \citep{huang2021model} & target & target & \checkmark & {\scalebox{0.7}{\XSolidBrush}} \\
    \method (ours) & p. target & source & \checkmark & \checkmark \\
    \bottomrule
    \end{tabular}
    \vspace{-10pt}
\end{table}

\begin{figure*}[t]
    \centering
    \includegraphics[width=1\textwidth]{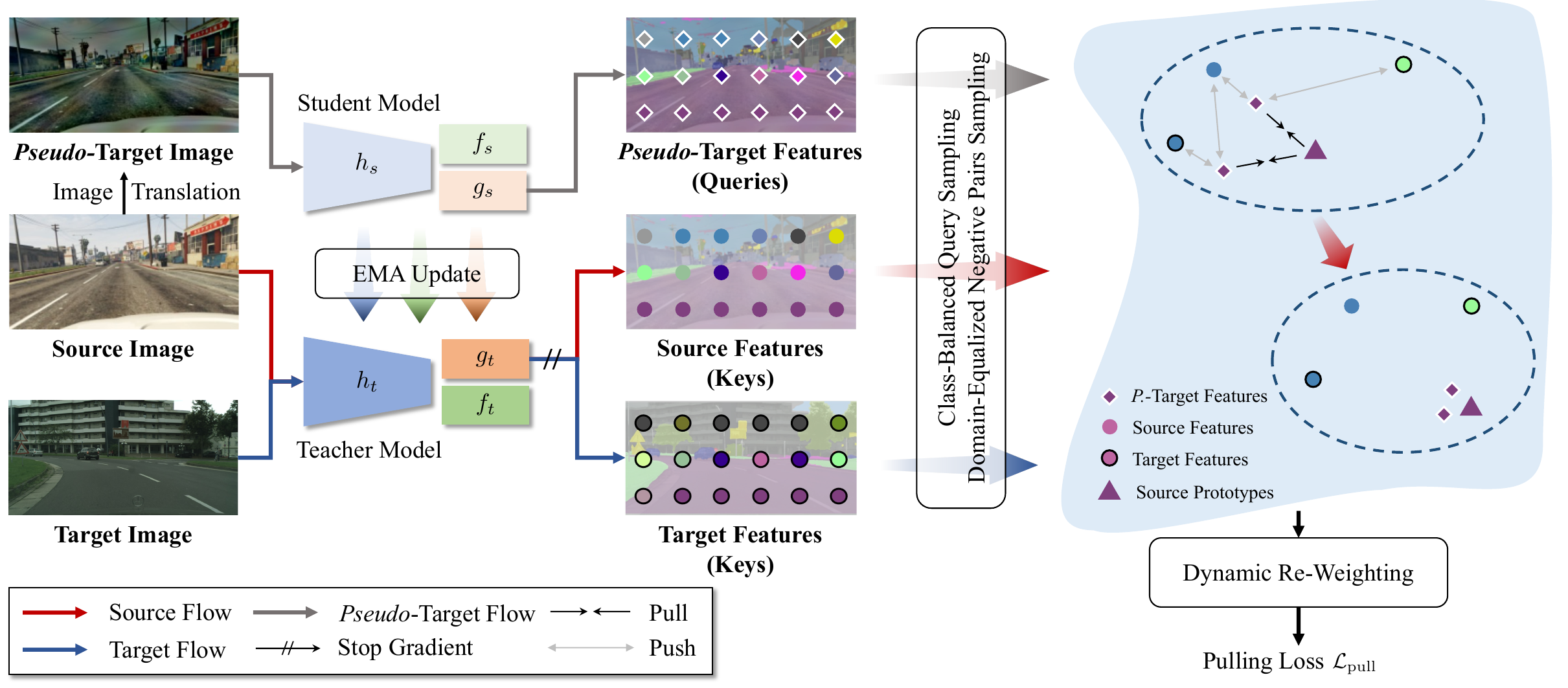}
    \caption{%
        \textbf{Illustration of contrastive pairs.}
        We regard features from \textit{pseudo-targets} and \textit{source prototypes} of the same category as the positive pairs defined in Eq. (\ref{eq:pos}).
        Negative keys include 1) source features from a different category (as in Eq. (\ref{eq:neg_s})), and 2) unreliable target features from a different category (as in Eq. (\ref{eq:neg_t})).
        In this way, this model is encouraged to \textit{learn similar features between the source and target domains from any category}.
        The improved similarity indeed boosts segmentation results (see Fig. \ref{fig:stats}).
        %
    }
    \label{fig:pipeline}
    \vspace{-10pt}
\end{figure*}

{\color{black}

\subsection{Discrepancy-based Domain Adaptation}

Most previous discrepancy-based methods align source and target domains via adversarial learning \citep{goodfellow2014generative}.
Specifically, \citep{saito2018maximum} maximizes and minimizes the discrepancy on target by introducing an adversarial objective in the second and third stages, respectively.
\citep{ganin2016domain} proposed domain-adversarial neural networks motivated by the theoretical analysis by \citep{ben2006analysis}.
\citep{tzeng2017adversarial} combined discriminative modeling, untied weight sharing, and a GAN loss.
\citep{bousmalis2017unsupervised} and \citep{hoffman2018cycada} further leveraged image translation models to minimize the domain discrepancy by cheating the discriminator.
The main difference between our \method and those GAN-based methods is that \method regards the labeled source domain as \textit{anchors} and \textit{explicitly aligns the joint feature distribution to the source feature space}, while those methods implicitly align cross-domain features to an \textit{intermediate common space}.
The advantage of pulling target to source compared with aligning to a common space is that our \method manages to keep the performance on the \textit{source} domain (Tab.~\ref{tab:source}), while GAN-based methods are expected to perform worse than the source-only baseline on the source domain.
This indicates that \method learns more robust features and has better generalization abilities, since \method improves the performance on \textit{both domains}.
\textcolor{black}{
Another line of work directly minimizes the distance between the feature distributions of different domains.
Specifically,
DAN~\citep{long2015learning} proposes an MK-MMD~\citep{gretton2012optimal}-based
multi-layer adaptation regularizer, which is activated in the last three linear layers of a CNN.
RTN~\citep{long2016unsupervised}, on the basis of DAN, assumes that source and target classifiers differ by a residual function, and applies extra classifier adaption via a newly proposed residual transfer module.
Deep CORAL~\citep{sun2016deep} extends CORAL~\citep{sun2016return}, which aligns the second-order statistics of the source
and target distributions, to deep models through an auxiliary loss function for cross-domain outputs.
Compared with these distance-based methods~\citep{sun2016deep, long2015learning, long2016unsupervised}, this work does not aim to explore a specific distance.
We demonstrate that as long as regarding source features as anchors and pulling target features close to source ones for each category, a simple MSE manages to be effective.
}

}

\section{Method}\label{sec:method}


\begin{figure*}[t]
    \centering
    \includegraphics[width=0.9\linewidth]{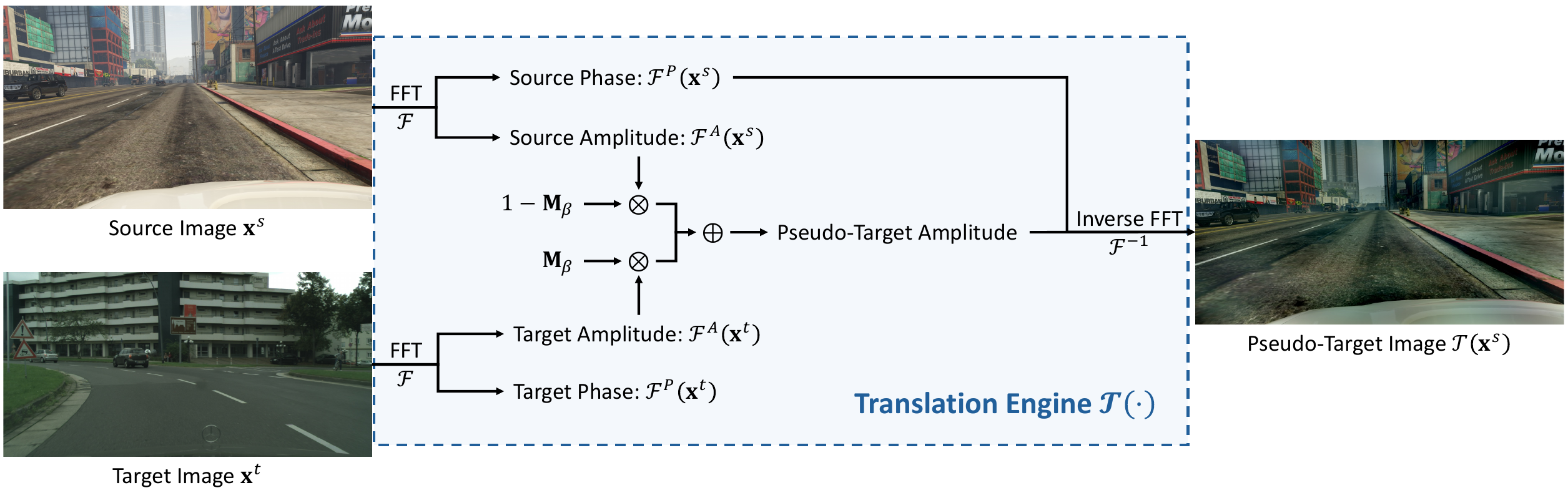}
    \vspace{-5pt}
    \caption{%
        \textbf{The pipeline of FDA}~\citep{yang2020fda}.
        Given a source image $\mathbf{x}^s$ and a randomly sampled target image $\mathbf{x}^t$, FDA transfers the source image into target style, resulting in $\mathcal{T}(\mathbf{x}^s)$, by pasting the low-frequency part of the amplitude from the target sample to the source image.
    }
    \label{fig:fda_pipeline}
    \vspace{-10pt}
\end{figure*}

\subsection{Overview}
\label{sec:overview}

Unsupervised domain adaptive semantic segmentation aims at applying a model learned from the labeled source dataset $\mathcal{D}_s=\left\{(\mathbf{x}_i^s, \mathbf{y}_i^s)\right\}_{i=1}^{n_s}$ to the unlabeled target dataset $\mathcal{D}_t=\left\{\mathbf{x}_i^t\right\}_{i=1}^{n_t}$,
where $n_s$ and $n_t$ are the numbers of images of source and target domains, respectively.
$\mathbf{x}_i^s\in\mathbb{R}^{H\times W\times 3}$ and $\mathbf{x}_i^t\in\mathbb{R}^{H\times W\times 3}$ are RGB images, while $\mathbf{y}_i^s\in\mathbb{R}^{H\times W\times C}$ is the one-hot semantic map associated with $\mathbf{x}_i^s$.
Source and target datasets share the same label space with $C$ categories.
Following previous methods \citep{hoyer2022daformer, zhang2021prototypical, li2022class, du2022learning, wang2023hard}, we adopt self-training as the baseline, where source data are used under a supervised manner and the unsupervised loss is computed based on target images and their pseudo-labels generated by a momentum teacher.
Concretely, our proposed \method follows the Mean Teacher \citep{meanteacher} framework that consists of a student and a teacher.
Each model $\psi$ reparameterized by $\theta$ consists of an encoder $h: \mathbb{R}^{H\times W\times 3} \to \mathbb{R}^{h\times w\times D}$ followed by a segmentor $f: \mathbb{R}^{h\times w\times D} \to (0,1)^{H\times W\times C}$ and a projector $g: \mathbb{R}^{h\times w\times D} \to \mathbb{R}^{h\times w\times d}$,
where $H$ and $W$ indicate the height and width of the input image respectively.
$h$, $w$ and $D$ form the shape of the intermediate features, and $d$ is the feature dimension of the projector.
The teacher is momentum updated by the student, \textit{i.e.}, $\theta_t \leftarrow \eta\theta_t + (1 - \eta)\theta_s$,
where $\eta$ is a momentum coefficient.
$\theta_s$ and $\theta_t$ are parameters of the student $\psi_s$ and the teacher $\psi_t$, respectively.

The overall objective is the sum of source supervised loss, target unsupervised loss, and pulling loss:
$\mathcal{L} = \mathcal{L}_{\mathrm{source}} + \mathcal{L}_{\mathrm{target}} + \lambda \mathcal{L}_{\mathrm{pull}}$.
For source data, we naively train the student with categorical cross-entropy (CE)
\begin{equation}
\label{eq:sup}
    \mathcal{L}_{\mathrm{source}} = \sum_{i=1}^{n_s} \sum_{j=1}^{H\times W} \ell_{ce}\left[f_s  (h_s(\mathbf{x}_{i}^s))(j) , \mathbf{y}_i^s(j)\right],
\end{equation}
where $\ell_{ce}(\mathbf{p}, \mathbf{y}) = - \mathbf{y}^\top\log\mathbf{p}$.
For unlabeled target data, we adopt self-training, which minimizes the weighted cross-entropy loss between predictions and pseudo-labels $\hat{\mathbf{y}}^t$ generated by the teacher.
\begin{equation}
\label{eq:unsup}
    \mathcal{L}_{\mathrm{target}} = \sum_{i=1}^{n_t} \sum_{j=1}^{H\times W} r_i^t \cdot \ell_{ce}\left[f_s(h_s(\mathbf{x}_{i}^t))(j) , \hat{\mathbf{y}}_i^t(j)\right],
\end{equation}
where $\hat{\mathbf{y}}_i^t(j)$ is the one-hot pseudo-label generated by the teacher for $i$-th target image at position $j$.
Moreover, following \citep{hoyer2022daformer} and \citep{tranheden2021dacs}, we use $r^t_i$ the ratio of pixels whose softmax probability exceeds a threshold $\delta_p$ to be the metric to measure the quality of pseudo-labels of the $i$-th target image:
\begin{equation}
    r^t_i = \frac{1}{H\times W} \sum_{j=1}^{H\times W} \mathbbm{1}[\max_c f_t \circ h_t (\mathbf{x}_i^t) > \delta_p],
\end{equation}
where $\mathbbm{1}[\cdot]$ is the indicator function and $\delta_p$ is set to $0.968$ following \citep{tranheden2021dacs}.

As discussed in Sec. \ref{sec:intro}, we observe that the model trained with source data is able to build a capable source feature space, but when we apply it to target domain, the features run into undesirable chaos.
\textit{The performance drop is highly related to the cross-domain feature dissimilarity} caused by the domain gap.
Intuitively, if we pull target features to source ones, it implicitly implies an adequate feature space for segmenting target samples.
To this end, we aim to conduct \textit{cross-domain positive pairs}, \textit{i.e.}, $(\mathbf{q}, \mathbf{k}^+)$, from the \textit{exactly} same category but different domains, and then maximize their agreement.
We study two different objectives, including InfoNCE \citep{infonce} and MSE.
InfoNCE pulls together positive pairs $(\mathbf{q}, \mathbf{k}^{+})$ and pushes away negative pairs $(\mathbf{q}, \mathbf{k}^{-})$.
\begin{equation}
\label{eq:contra}
    \mathcal{L}_{\mathrm{InfoNCE}} (\mathbf{q}, \mathbf{k}^{+}) = - 
    \log\left[
    \frac{
    e^{(\mathbf{q}^\top \mathbf{k}^{+} / \tau)}
    }
    {
    e^{(\mathbf{q}^\top \mathbf{k}^{+} / \tau)} +
    \sum\limits_{\mathbf{k}^{-} \in \mathcal{K}^{-}_{\mathbf{q}}}
    e^{(\mathbf{q}^\top \mathbf{k}^{-} / \tau)}
    } 
    \right],
\end{equation}
where $\mathbf{q}$, $\mathbf{k}^{+}$ and $\mathbf{k}^{-}$ are $\ell_2$-normalized features, which are outputs of the projector, \revise{indicating queries, positive keys, and negative keys, respectively.}
$\tau$ indicates the temperature.
$\mathcal{K}^{-}_{\mathbf{q}}$ is the set of negative keys of the given query $\mathbf{q}$, which is introduced in Sec. \ref{sec:contra}.
$\mathcal{L}_{\mathrm{MSE}} (\mathbf{q}, \mathbf{k}^{+}) = ||\mathbf{q} - \mathbf{k}^{+}||_2^2$, which maximizes the similarity of positive pairs $(\mathbf{q}, \mathbf{k}^{+})$ directly.

The pulling objective is \textit{weighted} over each positive pair
\begin{equation}
\mathcal{L}_{\mathrm{pull}} = \frac{1}{C} \sum_{c=0}^{C-1} w_c^{*} \sum_{(\mathbf{q}, \mathbf{k}^{+}) \in \mathcal{K}_c^{+}} \mathcal{L}_{\mathrm{pull}} (\mathbf{q}, \mathbf{k}^{+}),
\end{equation}
where $\mathcal{K}_c^{+}$ indicates the set of positive keys for category $c$ described later in Sec. \ref{sec:contra}. 
$w_c^{*}$ is the weight of class $c$, which is dynamically adjusted and discussed in Sec. \ref{sec:class_balanced}.
$\mathcal{L}_{\mathrm{pull}} (\mathbf{q}, \mathbf{k}^{+})$ is the pulling loss given a pair of positive features, which is either MSE or InfoNCE \citep{infonce}.

{\color{black}

\subsection{Preliminary: Fourier Domain Adaptation}
\label{sec:fda}

Fig.~\ref{fig:fda_pipeline} illustrates how Fourier domain adaptation (FDA)~\citep{yang2020fda} works.
In summary, given a source image $\mathbf{x}^s$ and a target image $\mathbf{x}^t$, FDA transfers the source sample into target style, \textit{i.e.}, $\mathcal{T}(\mathbf{x}^s)$, where $\mathcal{T}(\cdot)$ indicates the transformation function, \textit{e.g.}, FDA \citep{yang2020fda}.
Specifically, for every single channel image $\mathbf{x} \in \mathbb{R}^{H\times W}$:
\begin{equation}
    \mathcal{F}(\mathbf{x}) (m, n) = \sum_{h,w} \mathbf{x}(h, w) \exp\left[-j2\pi\left(\frac{h}{H} m + \frac{w}{W} n\right)\right],
\end{equation}
where $\mathcal{F}$ is the Fourier transform function of an RGB image, and $j^2 = -1$.
It can be efficiently implemented by the FFT algorithm~\citep{frigo1998fftw}.
Accordingly, $\mathcal{F}^{-1}$ is the inverse function, mapping spectral signals back to RGB image spaces.
Next,~\citep{yang2020fda} define a binary mask $\mathbf{M}_{\beta} \in \{0,1\}^{H\times W}$, whose value keeps zero except for the center region, according to a specific control value of $\beta \in (0,1)$.
Concretely, when we assume the center of the image is $(0, 0)$, the formulation becomes:
\begin{equation}
    \mathbf{M}_{\beta}(h,w) = \mathbbm{1}_{(h,w) \in [-\beta H:\beta H, -\beta W : \beta W]},
\end{equation}
where $\mathbbm{1}[\cdot]$ is the binary indicator.

Given a source image $\mathbf{x}^s$ and a target image $\mathbf{x}^t$, FDA transfers the source sample into target style by:
\begin{equation}
    \mathcal{T}(\mathbf{x}^s) = \mathcal{F}^{-1} [\mathbf{M}_{\beta} \odot \mathcal{F}^A (\mathbf{x}^t) + (1 - \mathbf{M}_{\beta}) \odot \mathcal{F}^A(\mathbf{x}^s), \mathcal{F}^P(\mathbf{x}^s)],
\end{equation}
where $\mathcal{F}^A, \mathcal{F}^P: \mathbb{R}^{H\times W\times 3} \to \mathbb{R}^{H\times W\times 3}$ be the amplitude and phase components of the Fourier transform $\mathcal{F}$ of an RGB image, respectively, and ``$\odot$'' is the element-wise dot production.
The underlying assumption is that the \textit{low-frequency} part of the amplitude contains domain-specific information.
Therefore, replacing the low-frequency part of $\mathcal{F}^A(\mathbf{x}^s)$ by that of the target image $\mathbf{x}^t$ is all we need to obtain $\mathcal{T}(\mathbf{x}^s)$.
Based on these operations, those pseudo-target images $\mathcal{T}(\mathbf{x}^s)$ should:
\begin{itemize}
    \item \textit{Guarantee semantic alignment at the pixel-level with $\mathbf{x}^s$}, which means for each pixel $(h,w)$, the ground-truth label of $\mathcal{T}(\mathbf{x}^s)(h,w)$ keeps the same with that of $\mathbf{x}^s(h,w)$. This is because the low-level spectrum (amplitude) can vary significantly \textit{without} affecting the perception of high-level semantics as demonstrated by~\citep{yang2020fda}.
    \item \textit{Appear in the style of target image $\mathbf{x}^t$}. This is because the low-level spectrum (amplitude) controls the appearance of an image according to~\citep{yang2020fda}.
\end{itemize}
These two characteristics together make it possible to assign precise positive samples $(\mathbf{q}, \mathbf{k}^{+})$, allowing \method to learn similar cross-domain features regarding each category.

}

\subsection{Pulling Target to Source}
\label{sec:contra}

In this section, we describe how to generate positive pairs that \textit{exactly} belong to the same category without target annotations.
Next, we take $\mathcal{L}_{\mathrm{InfoNCE}}$ as the alignment objective, and provide detailed formulations.

We employ an image translation engine to make sure cross-domain positive pairs belong to the same category.
Illustrated in Fig. \ref{fig:pipeline}, we first feed source data $\mathbf{x}^s$ into the image translation engine $\mathcal{T}$ (FDA \citep{yang2020fda} in this paper) to produce \textit{pseudo-target} data $\mathcal{T}(\mathbf{x}^s)$, which naturally have annotations, and then urge the model to learn similar features across source and pseudo-target domains regarding each category,
\textit{i.e.}, pulling features of $\mathcal{T}(\mathbf{x}^s)$ \textit{close} to features of $\mathbf{x}^s$ for each class.
Formulations are provided as follows.

\paragraph{Queries $\mathbf{q}$}
are features to be optimized, and thus they come from pseudo-target images $\mathcal{T}(\mathbf{x}^{s})$ 
\begin{equation}
\label{eq:query}
    \mathcal{Q}_c = \{ 
    g_s(h_s(\mathcal{T}(\mathbf{x}^s)))(j)
    \mid y^s(j,c) = 1
    \},
\end{equation}
where $j=1,2,\dots,h\times w$ is the pixel index, and we randomly sample $n_q(c)$ queries for class $c$ at each iteration, which will be discussed later in Sec. \ref{sec:sample}.

\paragraph{Positive pairs $(\mathbf{q}, \mathbf{k}^{+})$} are the crux for \method.
To learn similar cross-domain features, their agreements are supposed to be maximized.
Given a query $\mathbf{q} \in \mathcal{Q}_c$ belongs to class $c$, its positive key $\mathbf{k}^{+}$ is defined as the \textit{source prototype} of class $c$ generated by the momentum teacher.
Concretely, given an mini-batch $\mathcal{B} = \left( \mathbf{x}^s, \mathbf{y}^s, \mathbf{x}^t \right)$, we compute the source prototype of class $c$ by
\begin{equation}
\label{eq:pos}
    \mathbf{k}^{+}_c = \frac{
    \sum_{j=1}^{h\times w} \mathbbm{1}[y^s(j,c)=1] \cdot \left[g_t(h_t(\mathbf{x}^s))(j)\right]
    }{
    \sum_{j=1}^{h\times w} \mathbbm{1}[y^s(j,c)=1]
    }.
\end{equation}
Therefore, the set of positive pairs is defined as
\begin{equation}
    \mathcal{K}^{+} = \bigcup_{c=0}^{C-1} \left\{
    (\mathrm{norm}(\mathbf{q}), \mathrm{norm}(\mathbf{k}^{+}_c)) \mid 
    \mathbf{q}\in\mathcal{Q}_c
    \right\},
\end{equation}
where $\mathrm{norm}(\cdot)$ represents the $\ell_2$-norm, and $\mathcal{Q}_c$ is the set of candidate queries defined in Eq. (\ref{eq:query}).
Thanks to the image translation engine, we manage to conduct $\mathbf{q}$ and $\mathbf{k}^{+}$ that belong to the \textit{exact} category even without target labels.

\begin{figure}[t]
    \centering
    \includegraphics[width=1\linewidth]{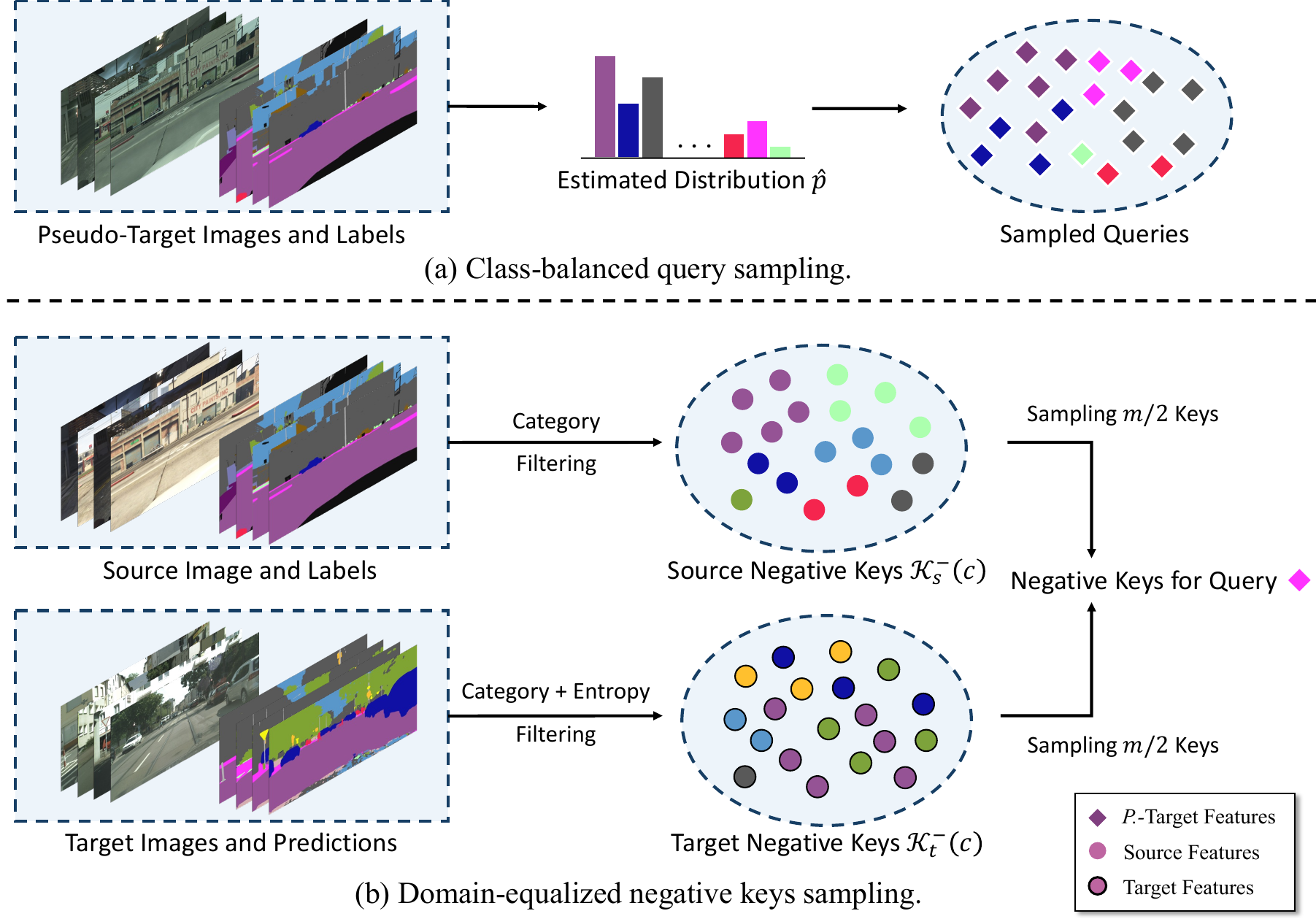}
    \caption{
    \textbf{Illustration of our sampling strategies} introduced in Sec~\ref{sec:sample}.
    (a) Class-balanced query sampling means we first estimate the current label distribution and then sample queries that follow the estimated distribution.
    (b) Domain-equalized negative keys sampling indicates we equally sample $m/2$ negative keys for each query from source and target, respectively.
    }
    \label{fig:sampling}
    \vspace{-10pt}
\end{figure}

\paragraph{Negative pairs $(\mathbf{q}, \mathbf{k}^{-})$} are used to avoid model collapse \citep{he2020momentum, chen2020simple} in InfoNCE \citep{infonce}, which is not out main focus.
Given a query $\mathbf{q} \in \mathcal{Q}_c$ belongs to class $c$, its negative keys consist of features from both domains.
For source features, we simply choose features that do not belong to category $c$
\begin{equation}
\label{eq:neg_s}
    \mathcal{K}_s^{-}(c) = \left\{
    g_t(h_t(\mathbf{x}^s))(j) \mid
    y^s(j,c) = 0
    \right\},
\end{equation}
where $j=1,2,\dots,h\times w$ denotes the pixel index.
The subscript $s$ here stands for ``source''.
For target features, inspired by \citep{wang2022semi} and \citep{wang2023using}, we take those unreliable features into consideration, fully mining the inherited information behind the target domain.
Specifically, we regard features (1) whose confidence is in the last $\alpha$ percent and (2) are predicted not to belong to category $c$ as candidate negative samples
\begin{equation}
\label{eq:neg_t}
\begin{aligned}
    \mathcal{K}_t^{-}(c) = \{
    g_t(h_t(\mathbf{x}^t))(j) \mid
    &\max_{c'} f_t(h_t (\mathbf{x}^t))(j, c') < \gamma,\\
    &\hat{y}^t(j,c)=0
    \},
\end{aligned}
\end{equation}
where $\max_{c'} f_t(h_t (\mathbf{x}^t))(j, c')$ indicates the confidence (\textit{i.e.}, maximum probability) of target image $\mathbf{x}^t$ at pixel $j$.
The subscript $t$ here stands for ``target''.
$\gamma$ is the confidence threshold related to unreliable partition $\alpha$, satisfying
\begin{equation}
\label{eq:alpha}
    \sum_{j=1}^{h \times w} \mathbbm{1}[\max_{c'} f_t(h_t (\mathbf{x}^t))(j, c') < \gamma] = \alpha \cdot (hw),
\end{equation}
\textit{i.e.}, \texttt{$\gamma$=np.percentile(C.flatten(),100*$\alpha$)}, and \texttt{C} is the confidence matrix.
Overall, the set of negative keys for query $\mathbf{q}$ is:
\begin{equation}
    \mathcal{K}_{\mathbf{q}}^{-} = \bigcup_{c=0}^{C-1} \left\{
    \mathrm{norm}(\mathbf{k}^{-}_c) \mid \mathbf{k}^{-}_c\in\mathcal{K}_s^{-}(c)\cup\mathcal{K}_t^{-}(c)
    \right\}.
\end{equation}

\paragraph{Discussion.}
%
%
\method tries to make target features (usually noisy) close to source features without target annotations.
Therefore, we regard pseudo-target features as queries and try to make them similar to \textit{consistent} positive keys (\textit{i.e.}, source features extracted by the momentum teacher).

\subsection{Sampling Strategies}
\label{sec:sample}

\noindent\textbf{Class-balanced query sampling.}
Semantic segmentation usually suffers from a long-tailed label distribution.
To alleviate this issue, a vanilla method is class-equalized query sampling (CEQS), treating all categories equally, \textit{i.e.}, the number of queries $n_q(c)$ of class $c$ equals to $n$ for each $c$.
However, CEQS tends to oversample rare classes whose features are usually noisy and unstable, leading to extra training noise.
To this end, we adopt class-balanced query sampling (CBQS) based on the label distribution of the current mini-batch.
Specifically, given an input mini-batch $\mathcal{B} = \left( \mathbf{x}^s, \mathbf{y}^s, \mathbf{x}^t \right)$, we first compute the distribution of $\mathbf{y}^s$
\begin{equation}
    \hat{p}(c) = \frac{
    \sum_{j=1}^{h\times w}\mathbbm{1}[y^s(j,c) = 1]
    }{
    \sum_{c=0}^{C-1}\sum_{j=1}^{h\times w}\mathbbm{1}[y^s(j,c) = 1]
    }.
\end{equation}
Next, based on $\hat{\mathbf{p}}$ and the base sample number $n$, we define $n_q(c) = \lceil C\cdot \hat{p}(c)\cdot n \rceil$.

\paragraph{Domain-equalized negative pair sampling.}
After we get the set of candidate negative pairs, a straightforward solution is to sample $m$ negative pairs from $\mathcal{K}^{-}$ directly.
However, the proportion of negative samples from each domain (\textit{i.e.}, source and target domains) can be unstable under such a strategy.
The more samples from $\mathcal{K}^{-}_s$ are introduced, the more homogeneous for all negative pairs.
By contrast, the more samples from $\mathcal{K}^{-}_t$ are applied, the more unstable training procedure we have due to the \textit{false negative} issue owing to the lack of target ground truths.
Therefore, to balance the stability during training (by introducing more $\mathbf{k}^{-}\in\mathcal{K}^{-}_s$) and the enrichment of negative samples (by introducing more $\mathbf{k}^{-}\in\mathcal{K}^{-}_t$), we adopt domain-equalized negative pair sampling (DENPS), where we randomly sample $m/2$ features from $\mathcal{K}^{-}_s$ and $m/2$ features from $\mathcal{K}^{-}_t$.

\subsection{Dynamic Re-Weighting}
\label{sec:class_balanced}

Due to the limited labeled source data and the existing domain shift, the training tends to be biased to dominant categories \citep{zou2018unsupervised, zou2019confidence}.
To address this, we propose a novel dynamic re-weighting strategy which urges the model paying more attention on underperforming classes and prevents it from being overwhelmed by well-trained samples.
Concretely, for class $c$, we define its weight $w^{*}_c$ in the pulling loss as
$w^{*}_c = w_c / \left(\sum_{c=0}^{C-1} w_c\right)$,
where $w_c$ is the weight computed based on the mean confidence
\begin{equation}
\label{eq:re-weighting}
    w_c = \left[\frac{1 - \mathrm{conf}(c)}{\max_{c'}(1 - \mathrm{conf}(c'))}\right]^{\beta},
\end{equation}
where $\beta=0.5$ is a scale factor and $\mathrm{conf}(c)$ denotes the mean confidence (\textit{i.e.}, maximum probability) of class $c$ on target in current mini-batch $\mathcal{B} = \left( \mathbf{x}^s, \mathbf{y}^s, \mathbf{x}^t \right)$, 
which can be jointly calculated by pseudo-labels $\hat{\mathbf{y}}^t$ and teacher predictions $f_t ( h_t (\mathbf{x}^t))$
\begin{equation}
    \mathrm{conf}(c) = \frac{\sum_{j=1}^{H\times W} \mathbbm{1}[\hat{y}^t(j, c) = 1] \cdot \left[\max_{c'} f_t(h_t (\mathbf{x}^t))(j, c')\right]}{\sum_{j=1}^{H\times W} \mathbbm{1}[\hat{y}^t(j, c) = 1]}.
\end{equation}

\section{Experiments}\label{sec:exp}

\begin{table*}[h]
\caption{
Comparison with state-of-the-art alternatives on \textbf{GTA5 $\to$ Cityscapes} benchmark with ResNet-101 \citep{he2016deep} and DeepLab-V2 \citep{chen2017deeplab}.
The results are averaged over 3 random seeds.
The top performance is highlighted in \textbf{bold} and the second score is \underline{underlined}.
}
\vspace{-5pt}
\centering\footnotesize
\setlength{\tabcolsep}{1.9pt}
\label{tab:gta_more}
\begin{tabular}{l | ccccccccccccccccccc | c}
\toprule
Method &
\rotatebox{90}{Road} & \rotatebox{90}{S.walk} & \rotatebox{90}{Build.} & \rotatebox{90}{Wall} & \rotatebox{90}{Fence} & \rotatebox{90}{Pole} & \rotatebox{90}{T.light} & \rotatebox{90}{Sign} & \rotatebox{90}{Veget.} & \rotatebox{90}{Terrain} & \rotatebox{90}{Sky} & \rotatebox{90}{Person} & \rotatebox{90}{Rider} & \rotatebox{90}{Car} & \rotatebox{90}{Truck} & \rotatebox{90}{Bus} & \rotatebox{90}{Train} & \rotatebox{90}{M.bike} & \rotatebox{90}{Bike} & mIoU \\
\midrule
source only & 70.2 & 14.6 & 71.3 & 24.1 & 15.3 & 25.5 & 32.1 & 13.5 & 82.9 & 25.1 & 78.0 & 56.2 & 33.3 & 76.3 & 26.6 & 29.8 & 12.3 & 28.5 & 18.0 & 38.6 \\
\midrule
AdaptSeg \citep{tsai2018learning} & 86.5 & 36.0 & 79.9 & 23.4 & 23.3 & 23.9 & 35.2 & 14.8 & 83.4 & 33.3 & 75.6 & 58.5 & 27.6 & 73.7 & 32.5 & 35.4 & 3.9 & 30.1 & 28.1 & 41.4 \\
CyCADA \citep{hoffman2018cycada} & 86.7 & 35.6 & 80.1 & 19.8 & 17.5 & 38.0 & 39.9 & 41.5 & 82.7 & 27.9 & 73.6 & 64.9 & 19.0 & 65.0 & 12.0 & 28.6 & 4.5 & 31.1 & 42.0 & 42.7 \\
ADVENT \citep{vu2019advent} & 89.4 & 33.1 & 81.0 & 26.6 & 26.8 & 27.2 & 33.5 & 24.7 & 83.9 & 36.7 & 78.8 & 58.7 & 30.5 & 84.8 & 38.5 & 44.5 & 1.7 & 31.6 & 32.4 & 45.5 \\
CBST \citep{zou2018unsupervised} & 91.8 & 53.5 & 80.5 & 32.7 & 21.0 & 34.0 & 28.9 & 20.4 & 83.9 & 34.2 & 80.9 & 53.1 & 24.0 & 82.7 & 30.3 & 35.9 & 16.0 & 25.9 & 42.8 & 45.9 \\
PCLA \citep{kang2020pixel} & 84.0 & 30.4 & 82.4 & 35.3 & 24.8 & 32.2 & 36.8 & 24.5 & 85.5 & 37.2 & 78.6 & 66.9 & 32.8 & 85.5 & 40.4 & 48.0 & 8.8 & 29.8 & 41.8 & 47.7 \\
FADA \citep{wang2020classes} & 92.5 & 47.5 & 85.1 & 37.6 & 32.8 & 33.4 & 33.8 & 18.4 & 85.3 & 37.7 & 83.5 & 63.2 & \underline{39.7} & 87.5 & 32.9 & 47.8 & 1.6 & 34.9 & 39.5 & 49.2 \\
MCS \citep{chung2022maximizing} & 92.6 & 54.0 & 85.4 & 35.0 & 26.0 & 32.4 & 41.2 & 29.7 & 85.1 & 40.9 & 85.4 & 62.6 & 34.7 & 85.7 & 35.6 & 50.8 & 2.4 & 31.0 & 34.0 & 49.7 \\
CAG \citep{zhang2019category} & 90.4 & 51.6 & 83.8 & 34.2 & 27.8 & 38.4 & 25.3 & 48.4 & 85.4 & 38.2 & 78.1 & 58.6 & 34.6 & 84.7 & 21.9 & 42.7 & \textbf{41.1} & 29.3 & 37.2 & 50.2 \\
FDA \citep{yang2020fda} & 92.5 & 53.3 & 82.4 & 26.5 & 27.6 & 36.4 & 40.6 & 38.9 & 82.3 & 39.8 & 78.0 & 62.6 & 34.4 & 84.9 & 34.1 & 53.1 & 16.9 & 27.7 & 46.4 & 50.5 \\
PIT \citep{lv2020cross} & 87.5 & 43.4 & 78.8 & 31.2 & 30.2 & 36.3 & 39.3 & 42.0 & 79.2 & 37.1 & 79.3 & 65.4 & 37.5 & 83.2 & \underline{46.0} & 45.6 & \underline{25.7} & 23.5 & 49.9 & 50.6 \\
IAST \citep{mei2020instance} & \underline{93.8} & 57.8 & 85.1 & 39.5 & 26.7 & 26.2 & 43.1 & 34.7 & 84.9 & 32.9 & 88.0 & 62.6 & 29.0 & 87.3 & 39.2 & 49.6 & 23.2 & 34.7 & 39.6 & 51.5 \\
DACS \citep{tranheden2021dacs} & 89.9 & 39.7 & \underline{87.9} & 30.7 & 39.5 & 38.5 & 46.4 & \underline{52.8} & \underline{88.0} & \underline{44.0} & \textbf{88.8} & 67.2 & 35.8 & 84.5 & 45.7 & 50.2 & 0.0 & 27.3 & 34.0 & 52.1 \\
RCCR \citep{zhou2021domain} & 93.7 & \underline{60.4} & 86.5 & 41.1 & 32.0 & 37.3 & 38.7 & 38.6 & 87.2 & 43.0 & 85.5 & 65.4 & 35.1 & \underline{88.3} & 41.8 & 51.6 & 0.0 & 38.0 & 52.1 & 53.5 \\  
ProDA \citep{zhang2021prototypical} & 91.5 & 52.4 & 82.9 & 42.0 & \underline{35.7} & 40.0 & 44.4 & 43.3 & 87.0 & 43.8 & 79.5 & 66.5 & 31.4 & 86.7 & 41.1 & 52.5 & 0.0 & 45.4 & \underline{53.8} & 53.7 \\  
CPSL \citep{li2022class} & 91.7 & 52.9 & 83.6 & \underline{43.0} & 32.3 & \textbf{43.7} & \underline{51.3} & 42.8 & 85.4 & 37.6 & 81.1 & \underline{69.5} & 30.0 & 88.1 & 44.1 & \textbf{59.9} & 24.9 & \underline{47.2} & 48.4 & \underline{55.7} \\
\method (ours) & \textbf{96.2} & \textbf{73.4} & \textbf{88.6} & \textbf{45.1} & \textbf{37.4} & \underline{40.7} & \textbf{54.0} & \textbf{55.5} & \textbf{88.9} & \textbf{48.6} & \underline{88.2} & \textbf{72.2} & \textbf{45.0} & \textbf{89.6} & \textbf{53.8} & \underline{56.2} & 1.3 & \textbf{53.0} & \textbf{59.6} & \textbf{60.4} \\
\midrule
\gc{target only$^\dag$} & \gc{97.1} & \gc{78.7} & \gc{90.3} & \gc{47.3} & \gc{52.3} & \gc{57.0} & \gc{62.9} & \gc{68.4} & \gc{91.7} & \gc{59.2} & \gc{93.8} & \gc{81.6} & \gc{60.5} & \gc{94.0} & \gc{65.6} & \gc{73.8} & \gc{59.3} & \gc{66.9} & \gc{77.0} & \gc{72.5} \\
\bottomrule
\end{tabular}
\end{table*}

\begin{table*}[t]
\caption{%
Comparison with state-of-the-art alternatives on \textbf{GTA5 $\to$ Cityscapes} benchmark using MiT-B5 \citep{xie2021segformer} as the backbone. 
The results are averaged over 3 random seeds.
The top performance is highlighted in \textbf{bold} font.
\dag\ means we reproduce the approach.
We incorporate our \method with three competitive baselines, \textit{i.e.}, DAFormer \citep{hoyer2022daformer}, HRDA \citep{hoyer2022hrda}, and MIC \citep{hoyer2023mic}.
}
\label{tab:gta}
\vspace{-5pt}
\centering\footnotesize
\setlength{\tabcolsep}{2pt}
\begin{tabular}{l | ccccccccccccccccccc | c}
\toprule
Method &
\rotatebox{90}{Road} & \rotatebox{90}{S.walk} & \rotatebox{90}{Build.} & \rotatebox{90}{Wall} & \rotatebox{90}{Fence} & \rotatebox{90}{Pole} & \rotatebox{90}{T.light} & \rotatebox{90}{Sign} & \rotatebox{90}{Veget.} & \rotatebox{90}{Terrain} & \rotatebox{90}{Sky} & \rotatebox{90}{Person} & \rotatebox{90}{Rider} & \rotatebox{90}{Car} & \rotatebox{90}{Truck} & \rotatebox{90}{Bus} & \rotatebox{90}{Train} & \rotatebox{90}{M.bike} & \rotatebox{90}{Bike} & mIoU \\
\midrule
source only$^\dag$
& 76.1 & 18.7   & 84.6    & 29.8 & 31.4   & 34.5 & 44.8  & 23.4 & 87.5 & 42.6 & 87.3	& 63.4   & 21.2  & 81.1 & 39.3 & 44.6 & 2.9 & 33.2 & 29.7 & 46.1 \\
DAFormer \citep{hoyer2022daformer}
& 95.7 & 70.2    & 89.4     & 53.5 & \textbf{48.1}  & 49.6 & 55.8  & 59.4 & 89.9       & 47.9   & 92.5 & 72.2   & 44.7  & 92.3 & 74.5  & \textbf{78.2} & 65.1   & 55.9      & 61.8 & 68.3 \\ 
DAFormer (w/ FDA)$^{\dag}$
& 95.4 & 68.2   & 89.8      & 52.9 & 45.1  & 51.4 & 60.9  & 51.2 & 90.1       & 48.6   & \textbf{92.6} &  \textbf{75.0}   & 45.9  & 93.0 & 72.4  & 74.3 & 62.1  & \textbf{62.3}       & 66.3 & 68.8 \\
\method (w/ DAFormer)
& \textbf{95.9} & \textbf{71.1}    & \textbf{89.9}     & \textbf{54.5} & 46.6  & \textbf{52.4} & \textbf{61.6}  & \textbf{61.0} & \textbf{90.3}       & \textbf{49.8}   & 92.5 & 74.6   & \textbf{46.3}  & \textbf{93.2} & \textbf{75.8}  & 76.3 & \textbf{72.4}   & 58.9      & \textbf{66.8} & \textbf{70.0} \\ 
\gc{target only$^\dag$} & \gc{98.9} & \gc{80.1} & \gc{93.1} & \gc{63.3} & \gc{60.7} & \gc{64.3} & \gc{68.9} & \gc{70.1} & \gc{93.2} & \gc{59.8}  & \gc{94.2} & \gc{82.1} & \gc{67.5} & \gc{94.1} & \gc{75.6} & \gc{83.8} & \gc{79.1} & \gc{68.5} & \gc{77.1} & \gc{77.6} \\
\midrule
HRDA \citep{hoyer2022hrda} &
96.4 & 74.4 & \textbf{91.0} & 61.6 & 51.5 & 57.1 & 63.9 & \textbf{69.3} & \textbf{91.3} & 48.4 & \textbf{94.2} & \textbf{79.0} & 52.9 & \textbf{93.9} & \textbf{84.1} & 85.7 & 75.9 & 63.9 & 67.5 & 73.8 \\
T2S-DA (w/ HRDA) &
\textbf{96.8} & \textbf{76.2} & 90.8 & \textbf{67.3} & \textbf{56.1} & \textbf{59.7} & \textbf{64.3} & 68.9 & 90.7 & \textbf{53.0} & 92.5 & 78.3 & \textbf{56.1} & 93.7 & 81.8 & \textbf{86.3} & \textbf{76.2} & \textbf{67.3} & \textbf{70.1} & \textbf{75.1} \\
\midrule
MIC \citep{hoyer2023mic} & 
97.4 & 80.1 & 91.7 & 61.2 & 56.9 & 59.7 & 66.0 & \textbf{71.3} & \textbf{91.7} & 51.4 & \textbf{94.3} & \textbf{79.8} & 56.1 & 94.6 & 85.4 & 90.4 & 80.4 & 64.5 & 68.5 & 75.9 \\
\method (w/ MIC) &
\textbf{97.5} & \textbf{80.3} & \textbf{91.8} & \textbf{64.2} & \textbf{60.1} & \textbf{60.2} & \textbf{66.3} & 70.9 & 91.5 & \textbf{54.1} & 93.6 & 79.1 & \textbf{59.3} & \textbf{94.7} & \textbf{85.8} & \textbf{90.7} & \textbf{80.9} & \textbf{66.1} & \textbf{70.4} & \textbf{76.7} \\
\bottomrule
\end{tabular}
\end{table*}

\begin{table*}[h]
\caption{
Comparison with state-of-the-art alternatives on \textbf{Synthia $\to$ Cityscapes} benchmark with ResNet-101 \citep{he2016deep} and DeepLab-V2 \citep{chen2017deeplab}. 
The results are averaged over 3 random seeds.
The mIoU and the mIoU* indicate we compute mean IoU over 16 and 13 categories, respectively.
The top performance is highlighted in \textbf{bold} font and the second score is \textit{underlined}.
}
\label{tab:Synthia_more}
\vspace{-5pt}
\centering\footnotesize
\setlength{\tabcolsep}{2.7pt}
\begin{tabular}{l | cccccccccccccccc | cc}
\toprule
Method &
\rotatebox{90}{Road} & \rotatebox{90}{S.walk} & \rotatebox{90}{Build.} & \rotatebox{90}{Wall*} & \rotatebox{90}{Fence*} & \rotatebox{90}{Pole*} & \rotatebox{90}{T.light} & \rotatebox{90}{Sign} & \rotatebox{90}{Veget.} & \rotatebox{90}{Sky} & \rotatebox{90}{Person} & \rotatebox{90}{Rider} & \rotatebox{90}{Car} & \rotatebox{90}{Bus} & \rotatebox{90}{M.bike} & \rotatebox{90}{Bike} & mIoU & mIoU* \\
\midrule
source only$^\dag$ & 55.6 & 23.8 & 74.6 & 9.2 & 0.2 & 24.4 & 6.1 & 12.1 & 74.8 & 79.0 & 55.3 & 19.1 & 39.6 & 23.3 & 13.7 & 25.0 & 33.5 & 38.6 \\
\midrule
AdaptSeg \citep{tsai2018learning} & 79.2 & 37.2 & 78.8 & - & - & - & 9.9 & 10.5 & 78.2 & 80.5 & 53.5 & 19.6 & 67.0 & 29.5 & 21.6 & 31.3 & - & 45.9 \\
ADVENT \citep{vu2019advent} & 85.6 & 42.2 & 79.7 & 8.7 & 0.4 & 25.9 & 5.4 & 8.1 & 80.4 & 84.1 & 57.9 & 23.8 & 73.3 & 36.4 & 14.2 & 33.0 & 41.2 & 48.0 \\
CBST \citep{zou2018unsupervised} & 68.0 & 29.9 & 76.3 & 10.8 & 1.4 & 33.9 & 22.8 & 29.5 & 77.6 & 78.3 & 60.6 & 28.3 & 81.6 & 23.5 & 18.8 & 39.8 & 42.6 & 48.9 \\
CAG \citep{zhang2019category} & 84.7 & 40.8 & 81.7 & 7.8 & 0.0 & 35.1 & 13.3 & 22.7 & 84.5 & 77.6 & 64.2 & 27.8 & 80.9 & 19.7 & 22.7 & 48.3 & 44.5 & 51.5\\
PIT \citep{lv2020cross} & 83.1 & 27.6 & 81.5 & 8.9 & 0.3 & 21.8 & 26.4 & 33.8 & 76.4 & 78.8 & 64.2 & 27.6 & 79.6 & 31.2 & 31.0 & 31.3 & 44.0 & 51.8 \\
FDA \citep{yang2020fda} & 79.3 & 35.0 & 73.2 & - & - & - & 19.9 & 24.0 & 61.7 & 82.6 & 61.4 & 31.1 & 83.9 & 40.8 & 38.4 & 51.1 & - & 52.5 \\
FADA \citep{wang2020classes} & 84.5 & 40.1 & 83.1 & 4.8 & 0.0 & 34.3 & 20.1 & 27.2 & 84.8 & 84.0 & 53.5 & 22.6 & 85.4 & 43.7 & 26.8 & 27.8 & 45.2 & 52.5 \\
MCS \citep{chung2022maximizing} & \underline{88.3} & \textbf{47.3} & 80.1 & - & - & - & 21.6 & 20.2 & 79.6 & 82.1 & 59.0 & 28.2 & 82.0 & 39.2 & 17.3 & 46.7 & - & 53.2 \\
PyCDA \citep{lian2019constructing} & 75.5 & 30.9 & 83.3 & 20.8 & 0.7 & 32.7 & 27.3 & 33.5 & 84.7 & 85.0 & 64.1 & 25.4 & 85.0 & 45.2 & 21.2 & 32.0 & 46.7 & 53.3 \\
PLCA \citep{kang2020pixel} & 82.6 & 29.0 & 81.0 & 11.2 & 0.2 & 33.6 & 24.9 & 18.3 & 82.8 & 82.3 & 62.1 & 26.5 & 85.6 & 48.9 & 26.8 & 52.2 & 46.8 & 54.0 \\
DACS \citep{tranheden2021dacs} & 80.6 & 25.1 & 81.9 & 21.5 & \underline{2.9} & 37.2 & 22.7 & 24.0 & 83.7 & \textbf{90.8} & 67.6 & \underline{38.3} & 82.9 & 38.9 & 28.5 & 47.6 & 48.3 & 54.8 \\
RCCR \citep{zhou2021domain} & 79.4 & 45.3 & 83.3 & - & - & - & 24.7 & 29.6 & 68.9 & 87.5 & 63.1 & 33.8 & 87.0 & 51.0 & 32.1 & 52.1 & - & 56.8 \\
IAST \citep{mei2020instance} & 81.9 & 41.5 & 83.3 & 17.7 & \textbf{4.6} & 32.3 & 30.9 & 28.8 & 83.4 & 85.0 & 65.5 & 30.8 & 86.5 & 38.2 & 33.1 & 52.7 & 49.8 & 57.0 \\
ProDA \citep{zhang2021prototypical} & 87.1 & 44.0 & 83.2 & \textbf{26.9} & 0.7 & 42.0 & 45.8 & \underline{34.2} & \underline{86.7} & 81.3 & 68.4 & 22.1 & \underline{87.7} & 50.0 & 31.4 & 38.6 & 51.9 & 58.5 \\
SAC \citep{araslanov2021self} & \textbf{89.3} & \underline{47.2} & \underline{85.5} & \underline{26.5} & 1.3 & 43.0 & 45.5 & 32.0 & \textbf{87.1} & \underline{89.3} & 63.6 & 25.4 & 86.9 & 35.6 & 30.4 & 53.0 & 52.6 & 59.3 \\
CPSL \citep{li2022class} & 87.3 & 44.4 & 83.8 & 25.0 & 0.4 & 42.9 & \underline{47.5} & 32.4 & 86.5 & 83.3 & \underline{69.6} & 29.1 & \textbf{89.4} & \underline{52.1} & \underline{42.6} & \underline{54.1} & \underline{54.4} & \underline{61.7} \\ 
\method (ours) & 81.2 & 38.3 & \textbf{86.0} & \underline{26.5} & 1.8 & \textbf{43.8} & \textbf{48.0} & \textbf{54.6} & 85.2 & 86.6 & \textbf{73.0} & \textbf{40.8} & 87.5 & \textbf{52.8} & \textbf{52.2} & \textbf{62.6} & \textbf{57.6} & \textbf{65.4} \\ 
\midrule
\gc{target only$^\dag$} & \gc{97.1} & \gc{78.7} & \gc{90.3} & \gc{47.3} & \gc{52.3} & \gc{57.0} & \gc{62.9} & \gc{68.4} & \gc{91.7} & \gc{93.8} & \gc{81.6} & \gc{60.5} & \gc{94.0} & \gc{73.8} & \gc{66.9} & \gc{77.0} & \gc{74.6} & \gc{79.7} \\
\bottomrule
\end{tabular}
\end{table*}

\begin{table*}[t]
\centering
\caption{%
Comparison with state-of-the-art alternatives on \textbf{Synthia $\to$ Cityscapes} benchmark using DAFormer \citep{hoyer2022daformer} with MiT-B5 \citep{xie2021segformer}. 
The results are averaged over 3 random seeds.
The mIoU and the mIoU* indicate we compute mean IoU over 16 and 13 categories, respectively.
The top performance is highlighted in \textbf{bold} font.
\dag\ means we reproduce the approach.
We incorporate our \method with three competitive baselines, including DAFormer \citep{hoyer2022daformer}, HRDA \citep{hoyer2022hrda}, and MIC \citep{hoyer2023mic}.
}
\label{tab:Synthia}
\vspace{-5pt}
\centering\footnotesize
\setlength{\tabcolsep}{2.7pt}
\begin{tabular}{l | cccccccccccccccc | cc}
\toprule
Method &
\rotatebox{90}{Road} & \rotatebox{90}{S.walk} & \rotatebox{90}{Build.} & \rotatebox{90}{Wall*} & \rotatebox{90}{Fence*} & \rotatebox{90}{Pole*} & \rotatebox{90}{T.light} & \rotatebox{90}{Sign} & \rotatebox{90}{Veget.} & \rotatebox{90}{Sky} & \rotatebox{90}{Person} & \rotatebox{90}{Rider} & \rotatebox{90}{Car} & \rotatebox{90}{Bus} & \rotatebox{90}{M.bike} & \rotatebox{90}{Bike} & mIoU & mIoU* \\
\midrule
source only$^\dag$
& 56.5 & 23.3   & 81.3    & 16.0 & 1.3   & 41.0 & 30.0  & 24.1 & 82.4 & 82.5	& 62.3   & 23.8  & 77.7 & 38.1 & 15.0 & 23.7 & 42.4 & 47.7 \\
DAFormer \citep{hoyer2022daformer}
& 84.5 & 40.7   & 88.4    & 41.5 & \textbf{6.5}   & 50.0 & 55.0  & 54.6 & 86.0 & 89.8    & 73.2   & 48.2  & 87.2 & 53.2 & 53.9 & 61.7 & 60.9 & 67.4 \\ 
DAFormer (w/ FDA)$^{\dag}$
& 76.9 & 32.6   & 88.2    & 41.1 & 5.2   & \textbf{54.1} & \textbf{61.3}  & \textbf{55.7} & 87.1 & 90.0 & \textbf{76.8}   & \textbf{48.7}  & 87.8 & 55.4 & \textbf{57.2} & \textbf{63.7} & 61.4 & 67.8 \\ 
\method (w/ DAFormer)
& 87.6 & 46.0   & \textbf{88.8}    & \textbf{43.7} & 6.4   & 53.3 & 59.1  & 54.8 & \textbf{87.5} & \textbf{91.1}    & 75.7   & 47.6  & \textbf{88.2} & \textbf{58.0} & 54.9 & 62.4 & \textbf{62.8} & \textbf{69.4} \\ 
\gc{target only$^\dag$} & \gc{98.9} & \gc{80.1} & \gc{93.1} & \gc{63.3} & \gc{60.7} & \gc{64.3} & \gc{68.9} & \gc{70.1} & \gc{93.2} & \gc{94.2} & \gc{82.1} & \gc{67.5} & \gc{94.1} & \gc{83.8} & \gc{68.5} & \gc{77.1} & \gc{78.7} & \gc{82.4} \\
\midrule
HRDA \citep{hoyer2022hrda} &
85.2 & 47.7 & \textbf{88.8} & 49.5 & 4.8 & 57.2 & 65.7 & 60.9 & \textbf{85.3} & 92.9 & \textbf{79.4} & 52.8 & 89.0 & 64.7 & 63.9 & 64.9 & 65.8 & 72.4 \\
T2S-DA (w/ HRDA) &
\textbf{85.7} & \textbf{50.3} & 88.5 & \textbf{50.1} & \textbf{9.7} & \textbf{61.7} & \textbf{67.1} & \textbf{62.3} & 84.7 & \textbf{93.0} & 77.9 & \textbf{56.1} & \textbf{89.3} & \textbf{68.2} & \textbf{65.7} & \textbf{70.3} & \textbf{67.5} & \textbf{73.8} \\
\midrule
MIC \citep{hoyer2023mic} & 
86.6 & 50.5 & 89.3 & 47.9 & 7.8 & 59.4 & 66.7 & 63.4 & 87.1 & \textbf{94.6} & 81.0 & 58.9 & \textbf{90.1} & 61.9 & 67.1 & 64.3 & 67.3 & 74.0 \\
T2S-DA (w/ MIC) &
\textbf{86.1} & \textbf{51.3} & \textbf{89.4} & \textbf{49.4} & \textbf{10.3} & \textbf{60.3} & \textbf{67.0} & \textbf{64.2} & \textbf{87.2} & 94.1 & \textbf{81.2} & \textbf{60.7} & 89.7 & \textbf{64.3} & \textbf{70.3} & \textbf{68.8} & \textbf{68.4} & \textbf{74.8} \\
\bottomrule
\end{tabular}
\end{table*}

\noindent\textbf{Datasets.} 
{
\color{black}
We use three publicly available datasets.
\\
\textit{Cityscapes} \citep{cordts2016cityscapes} is a dataset of real urban scenes taken from $50$ cities, which is regarded as the \textit{target} domain.
Several hundreds of thousands of frames were acquired from a moving vehicle during the span of several months, covering spring, and summer.
We use finely annotated images which consist of $2,975$ training images, $500$ validation images, and $1,525$ test images, with a resolution of $2048\times1024$. 
Each pixel of the image is divided into $19$ categories. 
We adopt training images as the unlabeled target domain and operate evaluations on its validation set.
\\
\textit{GTA5} \citep{richter2016gta} is a composite image dataset sharing $19$ classes with Cityscapes, which is considered as the \textit{source} domain.
The pixel-level semantic segmentation ground truth for $24,966$ city scene images are extracted from the physically-based rendered computer game ``Grand Theft Auto V'' automatically and the labeling process was completed in only $49$ hours. These synthetic images are used as source domain data for training.
\\
\textit{Synthia} \citep{ros2016synthiaa} is a SYNTHetic collection of Imagery and Annotations of urban scenarios, which consists of photo-realistic frames rendered from a virtual city and comes with precise pixel-level semantic annotations, which is considered as the \textit{source} domain.
We select its subset following common practices \citep{chen2019progressive, hoyer2022daformer}, which has $16$ common semantic annotations with Cityscapes. 
In total, $9,400$ images with the resolution $1280\times760$ from the Synthia dataset are used as source data.

}

\paragraph{Training.}
We first resize target images to $1024\times 512$ and source images from GTA5 \citep{richter2016gta} to $1280\times 720$.
Then, we random crop source images to $1024\times 512$ and adopt FDA \citep{yang2020fda} to build the pseudo-target domain.
Finally, we randomly crop them into $512\times512$ for further training.
We adopt AdamW \citep{loshchilov2017adamw} with betas $(0.9, 0.999)$, a learning rate of $6\times 10^{-5}$ for the encoder and $6\times 10^{-4}$ for the segmentor and the projector, a weight-decay of $0.01$, linear learning rate warmup schedule with $t_{\mathrm{warm}} = 1.5\mathrm{k}$.
The model is trained with a batch of two source images and two target images, for $40\mathrm{k}$ iterations.
In accordance with \citep{tranheden2021dacs}, we set $\eta=0.999$.
Temperature $\tau=0.2$, scale factor $\beta=0.5$, unreliable partition $\alpha=50\%$, base number of queries $n=128$, number of negative pairs per query $m=1024$.
All experiments are conducted on 1 Telsa A100 GPU based on PyTorch \citep{paszke2019pytorch} and mmsegmentation \citep{mmseg2020}.

\paragraph{Network architecture.}
Start from using the DeepLab-V2 \citep{chen2017deeplab} as the segmentor $f$ with ResNet-101 \citep{he2016deep} as the encoder $h$, we adopt DACS \citep{tranheden2021dacs} with two extra strategies in \citep{hoyer2022daformer} as the baseline.
To further verify the efficiency of \method on recent Transformer-based networks, we use DAFormer \citep{hoyer2022daformer} as a stronger baseline, where $h$ is the MiT-B5 encoder \citep{xie2021segformer} and $f$ is the DAFormer decoder \citep{hoyer2022daformer}.
The projector $g$ takes the feature provided by the encoder as input and consists of two \texttt{Conv-BN-ReLU} blocks.

\subsection{Comparison with Existing Alternatives}
\label{sec:sota}

We compare our \method with a large selection of previous UDA methods.
The comparison is conducted on GTA5 \citep{richter2016gta} $\to$ Cityscapes \citep{zhou2017scene} using CNN-based models and Transformer-based models in Tab. \ref{tab:gta_more} and Tab. \ref{tab:gta}, respectively.
Moreover, the comparison is conducted on Synthia \citep{ros2016synthiaa} $\to$ Cityscapes \citep{zhou2017scene} using CNN-based models and Transformer-based models in Tab. \ref{tab:Synthia_more} and Tab. \ref{tab:Synthia}, respectively.

Selected methods can be summarized into three mainstream: (1) adversarial training based
methods for domain alignment, including AdaptSeg \citep{tsai2018learning}, CyCADA \citep{hoffman2018cycada},  ADVENT \citep{vu2019advent}, FADA \citep{wang2020classes}, 
(2) self-training based approaches, including CBST \citep{zou2018unsupervised}, IAST \citep{mei2020instance}, CAG \citep{zhang2019category}, ProDA \citep{zhang2021prototypical}, SAC \citep{araslanov2021self}, CPSL \citep{li2022class},
and (3) contrastive learning-based alternatives, including PLCA \citep{kang2020pixel}, RCCR \citep{zhou2021domain}, and MCS \citep{chung2022maximizing}.
Note that, we report the performances of ProDA \citep{zhang2021prototypical} and CPSL \citep{li2022class} in Tab. \ref{tab:gta_more} and Tab. \ref{tab:Synthia_more} without knowledge distillation using self-supervised trained models for a fair comparison.

As illustrated in Tab. \ref{tab:gta_more} and Tab. \ref{tab:Synthia_more}, when taking DeepLab-V2 \citep{chen2017deeplab} as the segmentation decoder and ResNet-101 \citep{he2016deep} as the backbone, our \method outperforms other competitors by a large margin, achieving mIoU of $60.4\%$ on GTA5 $\to$ Cityscapes, and $57.6\%$ over 16 classes and $65.4\%$ over 13 classes on Synthia $\to$ Cityscapes, respectively.
Especially for the class ``sidewalk'' on GTA $\to$ Cityscapes and the class ``bike'' on Synthia $\to$ Cityscapes, \method achieves IoU scores of $73.4\%$ and $62.6\%$, outperforming the second place by $+13.0\%$ and $+8.5\%$, respectively.
Moreover, significant improvements are observed when comparing our \method with other contrastive learning-based methods PLCA \citep{kang2020pixel}, MCS \citep{chung2022maximizing}, and RCCR \citep{zhou2021domain}.
%

\begin{figure*}[t]
    \centering
    \includegraphics[width=1\linewidth]{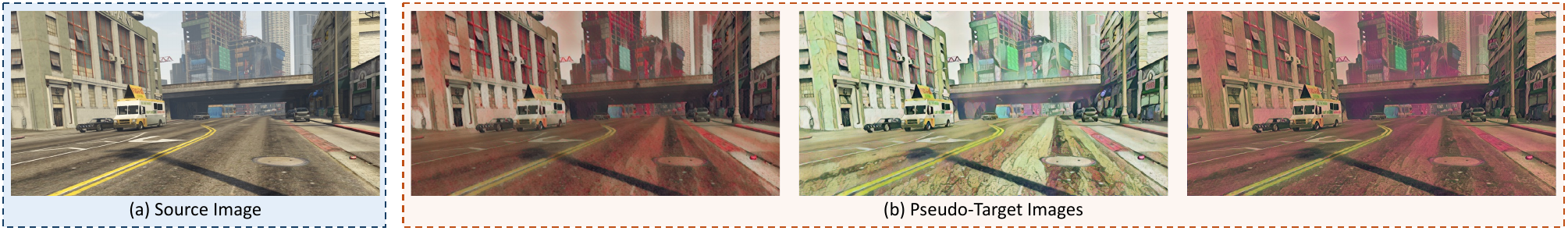}
    \vspace{-13pt}
    \caption{
    Illustration of the style augmentor proposed in \citep{kundu2021generalize}.
    We randomly augment the source image 3 times.
    We leverage this off-the-shelf augmentor as the image translation engine for domain generalization.
    }
    \label{fig:style}
    \vspace{-10pt}
\end{figure*}

More importantly, when equipped with an advanced Transformer \citep{dosovitskiy2020image}-based network, \method surpasses DAFormer \citep{hoyer2022daformer} by $+1.7\%$ mIoU on GTA5 $\to$ Cityscapes, $+1.9\%$ mIoU on Synthia $\to$ City-scapes (16 classes), and $+2.0\%$ mIoU* on Synthia $\to$ City-scapes (13 classes), as illustrated in Tab. \ref{tab:gta} and Tab. \ref{tab:Synthia}, respectively.
Benefiting from learning similar features across domains, \method improves the performances of class ``train'' by $+10.3\%$ compared to DAFormer \citep{hoyer2022daformer} on GTA5 $\to$ Cityscapes benchmark.
From these four tables, we can tell that by making the model learn similar features across domains, \method outperforms state-of-the-art competitors on various benchmarks.

{\color{black}

When incorporating with more advanced baselines, \textit{e.g.}, HRDA \citep{hoyer2022hrda} and MIC \citep{hoyer2023mic}, our \method brings significant improvements \textit{consistently} since our method is totally orthogonal to these two methods.

\paragraph{Implementation details.}
We simply implement our \method on the LR branch of HRDA \citep{hoyer2022hrda} with no modification.
Therefore, on the basis of HRDA, the objective becomes to $\mathcal{L} = \mathcal{L}_{\mathrm{LR}} + \mathcal{L}_{\mathrm{HR}} + \lambda\mathcal{L}_{\mathrm{LR}}^{\mathrm{pull}}$.
We did not implement $\mathcal{L}_{\mathrm{pull}}$ on the HR branch because we found it requires much more computational resources and GPU memory.
Similarly, we also implement our \method on the LR branch of MIC \citep{hoyer2023mic} as MIC was originally developed over HRDA.
}

\subsection{Domain Generalized Semantic Segmentation}
\label{sec:dg}
We further evaluate our proposed \method on domain generalized (DG) semantic segmentation, where images from the target domain are unaccessible, making it more important to extract domain-invariant feature representations.
We adopt the translation model provided by \citep{kundu2021generalize}.

{\color{black}
Specifically, there are five types of transformation functions by default.
(1) \textit{Domain-A}: Using FDA \citep{yang2020fda} while randomly selecting the target reference from a subset of style transfer dataset \citep{huang2017arbitrary}.
(2) \textit{Domain-B}: Leveraging a style transfer model \citep{jackson2019style} via randomly sampling a style embedding from a pre-defined multivariate normal distribution.
(3) \textit{Domain-C}: Using Adaptive Instance Normalization \citep{huang2017arbitrary} layers to inject style from a given reference.
The reference image is randomly sampled from the style transfer dataset \citep{huang2017arbitrary}.
(4) \textit{Domain-D}: Using realistic weather augmentations proposed by \citep{michaelis2019benchmarking}.
(5) \textit{Domain-E}: Using the technique proposed by \citep{imgaug} to generate cartoonized versions of input images.
Fig.~\ref{fig:style} illustrates the style augmentation, where we randomly augment the same source image 3 times by leveraging the off-the-shelf augmentor.
Implementation details can be found here\footnote{\url{https://github.com/Haochen-Wang409/StyleTransfer}}.
}

\begin{table}[t]
    \centering\small
    \caption{
    Comparison with previous methods on \textbf{domain generalized semantic segmentation}.
    All methods are trained on GTA5 and validated on Cityscapes.
    $\dag$ indicates our reproduced results.
    We implement SHADE \citep{zhao2022style} using the network architecture of DAFormer \citep{hoyer2022daformer}.
    }
    \label{tab:dg}
    \vspace{-5pt}
    \setlength{\tabcolsep}{4pt}
    \begin{tabular}{llcc}
    \toprule
    Method & Venue & Backbone & mIoU \\
    \midrule
    IBN-Net \citep{pan2018two} & ECCV'18 & \multirow{7}{*}{ResNet-101} & 37.4 \\
    DRPC \citep{yue2019domain} & ICCV'19 & & 42.5 \\
    ISW \citep{choi2021robustnet} & CVPR'21 & & 37.2 \\
    FSDR \citep{huang2021fsdr} & CVPR'21 & & 44.8 \\
    SAN-SAW \citep{peng2022semantic} & CVPR'22 & & 45.3 \\
    SHADE \citep{zhao2022style} & ECCV'22 & & 46.7 \\
    \midrule
    DAFormer$^{\dag}$ \citep{hoyer2022daformer} & CVPR'22 & \multirow{3}{*}{MiT-B5} & 46.1 \\
    SHADE$^{\dag}$ \citep{zhao2022style} & ECCV'22 & & 46.9 \\
    T2S-DA (w/ DAFormer) & Ours & & \textbf{48.6} \\
    \bottomrule
    \end{tabular}
\end{table}

\begin{table}[t]
    \centering
    \caption{
    Ablation studies on \textbf{designs of contrastive terms.}
    ``CE'', ``$\mathbf{q}$'', and ``$\mathbf{k}^{+}$'' are cross-entropy loss, queries, and positive keys, respectively.
    All experiments are conducted on the GTA5 $\to$ Cityscapes benchmark using Transformer-based models, \textit{i.e.}, DAformer \citep{hoyer2022daformer} with MiT-B5 \citep{xie2021segformer}.
    }
    \vspace{-5pt}
    \label{tab:abla_contra}
    \begin{tabular}{llllc}
    \toprule
    case & CE & $\mathbf{q}$ & $\mathbf{k}^{+}$ & mIoU \\
    \midrule
    DAFormer & source & - & - & 68.3{\scriptsize$\pm$0.5} \\
    DAFormer (w/ FDA) & p. target & - & - & 68.8{\scriptsize$\pm$0.8} \\
    vanilla contrast & source & source & source & 68.5{\scriptsize$\pm$0.3} \\
    source $\to$ target & source & source & target & 68.3{\scriptsize$\pm$1.1} \\
    source $\to$ p. target & source & source & p. target & 69.0{\scriptsize$\pm$0.3} \\
    T2S-DA (ours) & source & p. target & source & \textbf{70.0}{\scriptsize$\pm$0.6} \\
    p. target $\to$ source & p. target & p. target & source & 69.0{\scriptsize$\pm$1.6} \\
    target $\to$ source & source & target & source & 68.6{\scriptsize$\pm$1.3} \\
    \bottomrule
    \end{tabular}
    \vspace{-10pt}
\end{table}

Note that we cannot apply $\mathcal{L}_{\mathrm{target}}$, dynamic re-weighting, and domain-equalized negative pair sampling since we fail to get target samples in DG, and thus the overall objective becomes $\mathcal{L}_{\mathrm{source}} + \lambda \mathcal{L}_{\mathrm{pull}}$.

We take DAFormer \citep{hoyer2022daformer} trained with only source images in a fully-supervised manner as the baseline.
\revise{As illustrated in Tab. \ref{tab:dg}, when comparing with other alternatives, \method manages to bring significant improvements, verifying its domain-invariant property.}

\subsection{Ablation Studies}
\label{sec:abla}

In this section, we first the effectiveness of pulling target to source.
Specifically, we make various permutations of where to sample (1) queries and (2) their positive keys, and where the (3) cross-entropy loss should be applied.
Next, we study the effectiveness of two proposed sampling strategies and dynamic re-weighting.
Then, we study the effectiveness of different objectives by replacing InfoNCE with a simple MSE to verify our motivation is \textit{not} limited to contrastive learning. 
Finally, we study the effectiveness of different image translation engines and different hyper-parameters.
By default, all ablations in this section are conducted on the GTA5 $\to$ Cityscapes benchmark.
%

\begin{table}[t]
    \centering
    \setlength{\tabcolsep}{8pt}
    \caption{
    Ablation studies on \textbf{sampling strategies}, where ``CBQS'' indicates class-balanced query sampling and ``DENPS'' means domain-equalized negative pairs sampling.
    Significant improvement is observed only when combining these two techniques.
    }
    \label{tab:abla_sampling}
    \vspace{-5pt}
    \begin{tabular}{ccl}
    \toprule
    CBQS & DENPS & mIoU \\
    \midrule
    & & 68.7{\scriptsize$\pm$0.4} \\
    \checkmark & & 68.9{\scriptsize$\pm$0.3} \up{0.2} \\
    & \checkmark & 69.0{\scriptsize$\pm$0.4} \up{0.3} \\
    \checkmark & \checkmark & \textbf{70.0}{\scriptsize$\pm$0.6} \textbf{\up{1.3}} \\
    \bottomrule
    \end{tabular}
\end{table}

\begin{table}[t]
    \centering
    \setlength{\tabcolsep}{10pt}
    \caption{
    Ablation studies on \textbf{dynamic re-weighting} (DRW).
    Only when applying DRW to the alignment loss $\mathcal{L}_{\mathrm{pull}}$ brings improvements, especially in tailed classes.
    }
    \label{tab:abla_drw}
    \vspace{-5pt}
    \begin{tabular}{lll}
    \toprule
    DRW & mIoU & mIoU (Tail) \\
    \midrule
    none & 68.9{\scriptsize$\pm$0.6} & 62.0{\scriptsize$\pm$1.1} \\
    $\mathcal{L}_{\mathrm{target}}$ & 68.8{\scriptsize$\pm$0.4} \down{0.1} & 60.1{\scriptsize$\pm${1.3}} \down{1.8} \\
    $\mathcal{L}_{\mathrm{pull}}$ & \textbf{70.0}{\scriptsize$\pm$0.6} \textbf{\up{1.1}} & \textbf{63.7}{\scriptsize$\pm$0.8} \textbf{\up{1.7}} \\
    \bottomrule
    \end{tabular}
    \vspace{-10pt}
\end{table}

\begin{table}[t]
    \centering
    \setlength{\tabcolsep}{9pt}
    \caption{
    Ablation studies on \textbf{different objectives}.
    When replacing the advanced InfoNCE \citep{zou2018unsupervised} with a simple MSE, \method still brings significant improvements over baselines.
    }
    \label{tab:abla_loss}
    \vspace{-5pt}
    \begin{tabular}{llll}
    \toprule
    encoder & decoder & alignment & mIoU \\
    \midrule
    ResNet-101 & DeepLab-V2 & none & 56.3 \\
    ResNet-101 & DeepLab-V2 & MSE & 58.6 \up{2.3} \\
    ResNet-101 & DeepLab-V2 & InfoNCE & \textbf{60.4 \up{4.1}} \\
    \midrule
    MiT-B5 & DAFormer & none & 68.3 \\
    MiT-B5 & DAFormer & MSE & 69.4 \up{1.1} \\
    MiT-B5 & DAFormer & InfoNCE & \textbf{70.0 \up{1.7}} \\
    \bottomrule
    \end{tabular}
\end{table}

\begin{table}[t]
    \centering
    \caption{
    Ablations of using \textbf{different image translation models}, including FDA \citep{yang2020fda}, simple ColorJitter, and GaussianBlur.
    Also, two GAN-based methods, \textit{i.e.}, CyCADA \citep{hoffman2018cycada}, and \citep{wang2021uncertainty} are also studied.
    All methods are based on DAFormer \citep{hoyer2022daformer}.
    ``Training'' indicates whether the translation engine should be trained or not.
    ``$\mathcal{L}_{\mathrm{pull}}$'' means our T2S-DA objective.
    Without $\mathcal{L}_{\mathrm{pull}}$ implies that the source cross-entropy loss of DAFormer is applied on the pseudo-target domain.
    Significant improvements are observed consistently when we apply our $\mathcal{L}_{\mathrm{pull}}$ with \textit{various} kinds of translation models.
    }
    \label{tab:abla_translation}
    \vspace{-3pt}
    \setlength{\tabcolsep}{7pt}
    \begin{tabular}{lccl}
    \toprule
    Translation & Training & $\mathcal{L}_{\mathrm{pull}}$ & mIoU \\
    \midrule
    - & - & - & 68.3{\scriptsize{$\pm$}0.5} \\
    \citep{yang2020fda} & - & - & 68.8{\scriptsize{$\pm$}0.5} \\
    \citep{yang2020fda} & - & \checkmark & \textbf{70.0}{\scriptsize{$\pm$}0.6} \textbf{\up{1.2}} \\
    ColorJitter & - & - & 68.1{\scriptsize{$\pm$}0.9} \\
    ColorJitter & - & \checkmark & \textbf{69.0}{\scriptsize{$\pm$}0.6} \textbf{\up{0.9}} \\
    GaussianBlur & - & - & 68.4{\scriptsize{$\pm$}0.7} \\
    GaussianBlur & - & \checkmark & \textbf{69.5}{\scriptsize{$\pm$}0.5} \textbf{\up{1.1}} \\
    \citep{hoffman2018cycada} & \checkmark & - & 68.2{\scriptsize{$\pm$}1.3} \\
    \citep{hoffman2018cycada} & \checkmark & \checkmark & \textbf{69.2}{\scriptsize{$\pm$}0.7} \textbf{\up{1.0}} \\
    \citep{wang2021uncertainty} & \checkmark & - & 69.3{\scriptsize{$\pm$}0.6} \\
    \citep{wang2021uncertainty} & \checkmark & \checkmark & \textbf{70.7}{\scriptsize{$\pm$}0.3} \textbf{\up{1.4}} \\
    \bottomrule
    \end{tabular}
    \vspace{-10pt}
\end{table}

\begin{table*}[t]
\begin{minipage}{0.15\linewidth}
\centering
\caption{
Contrastive loss weight $\lambda$.
}
\vspace{4pt}
\label{tab:lambda}
\begin{tabular}{cc} 
\toprule
$\lambda$ & mIoU \\
\midrule
0.01 & 68.3{\scriptsize$\pm$0.9} \\
\textbf{0.1} & \textbf{70.0}{\scriptsize$\pm$0.6} \\
0.5 & 68.2{\scriptsize$\pm$0.6} \\
1 & 67.3{\scriptsize$\pm$0.4} \\
\bottomrule
\end{tabular}
\end{minipage}
\hfill
\begin{minipage}{0.15\linewidth}
\centering
\caption{
Temperature parameter $\tau$.
}
\vspace{4pt}
\label{tab:temp}
\begin{tabular}{cc} 
\toprule
$\tau$ & mIoU \\
\midrule
0.02 & 69.3{\scriptsize$\pm$0.4} \\
\textbf{0.2} & \textbf{70.0}{\scriptsize$\pm$0.6} \\
0.5 & 68.8{\scriptsize$\pm$0.7} \\
1 & 69.0{\scriptsize$\pm$0.1} \\
\bottomrule
\end{tabular} 
\end{minipage}
\hfill
\begin{minipage}{0.15\linewidth}
\centering
\caption{
Reweighting coefficient $\beta$.
}
\label{tab:beta}
\vspace{4pt}
\begin{tabular}{cc} 
\toprule
$\beta$ & mIoU \\
\midrule
0.1 & 69.9{\scriptsize$\pm$0.6} \\
\textbf{0.5} & \textbf{70.0}{\scriptsize$\pm$0.6} \\
1 & 68.8{\scriptsize$\pm$0.4}\\
2 & 69.0{\scriptsize$\pm$0.5} \\
\bottomrule
\end{tabular} 
\end{minipage}
\hfill
\begin{minipage}{0.15\linewidth}
\centering
\caption{
Unreliable partition $\alpha$.
}
\label{tab:apha}
\vspace{14pt}
\setlength{\tabcolsep}{6pt}
\begin{tabular}{cc} 
\toprule
$\alpha$ & mIoU \\
\midrule
10\% & 69.5{\scriptsize$\pm$0.1} \\
\textbf{50}\% & \textbf{70.0}{\scriptsize$\pm$0.6} \\
90\% & 69.4{\scriptsize$\pm$0.5}\\
\bottomrule
\end{tabular} 
\end{minipage}
\hfill
\begin{minipage}{0.31\linewidth}
\centering
\caption{
The number of queries $n$ and the number of negative pairs $m$.
}
\label{tab:n_m}
\vspace{-5pt}
\setlength{\tabcolsep}{15pt}
\begin{tabular}{ccc} 
\toprule
$n$ & $m$ & mIoU \\
\midrule
64 & 1024 & 68.7{\scriptsize$\pm$0.3} \\
\textbf{128} & \textbf{1024} & \textbf{70.0}{\scriptsize$\pm$0.6} \\
256 & 1024 & 69.3{\scriptsize$\pm$0.5} \\
128 & 512 & 69.5{\scriptsize$\pm$0.5} \\
128 & 2048 & 68.8{\scriptsize$\pm$0.3} \\
\bottomrule
\end{tabular} 
\end{minipage}
\vspace{-10pt}
\end{table*}

\paragraph{The effectiveness of pulling target to source} is studied in Tab.~\ref{tab:abla_contra},
where we take DAFormer \citep{hoyer2022daformer} as our baseline.
Simply apply FDA \citep{yang2020fda} to DAFormer \citep{hoyer2022daformer}, \textit{i.e.}, compute the supervised loss defined in Eq. (\ref{eq:sup}) on the pseudo-target domain, brings little improvements.
Vanilla contrast on the source domain also brings neglectable improvements.
When simply pulling source close to target, the performance maintains $68.3\%$ but with a larger standard deviation of $1.1\%$.
This is because, without target labels, it is hard to conduct positive pairs precisely.
To this end, we introduce an image translation engine to ensure true positive matches, but the improvements still remain limited, \textit{i.e.}, $69.0\%$.
However, when changing the direction of this alignment, \textit{i.e.}, \textit{pulling target to source}, however, the performance achieves $70.0\%$, bringing an improvement of $+1.7\%$.
In addition, if we compute cross-entropy on the pseudo target domain on the basis of \method, the performance drops to $69.0\%$.
\revise{
When simply pulling target features, the performance drops to $68.6\%$.
The advantage of using pseudo-target features over simple target features is that the former \textit{guarantees} the positive pairs $(\mathbf{q}, \mathbf{k}^{+})$ belong to the \textit{same} category, which is almost impossible when directly using target features due to the absence of target labels.
}
In conclusion, the alignment works \textit{only when we pull target to source leveraging the image translation engine}.
\textcolor{black}{
An intuitive explanation is provided below.
The alignment objective pulls a query $\mathbf{q}$ to its corresponding positive key $\mathbf{k}^{+}$, which means it regards $\mathbf{k}^{+}$ as an \textit{anchor} and tries to make $\mathbf{q}$ close to $\mathbf{k}^{+}$.
Therefore, where to sample $\mathbf{q}$ and its corresponding $\mathbf{k}^{+}$ determines the effectiveness of the alignment.
The selection of a pair of positive samples $(\mathbf{q}, \mathbf{k}^{+})$ should follow these rules:
\begin{itemize}
    \vspace{-6pt}
    \item $\mathbf{k}^{+}$ is expected to be categorically discriminative.
    \item $\mathbf{q}$ should (1) belong to a \textit{different domain} than $\mathbf{k}^{+}$, and (2) belong to the \textit{same category} as $\mathbf{k}^{+}$.
    \vspace{-6pt}
\end{itemize}
Therefore, 
\begin{itemize}
\vspace{-6pt}
\item \textit{Sampling $\mathbf{k}^{+}$ from the source domain} is the best choice because we have sufficient supervision signals on the source domain.
Otherwise, when $\mathbf{k}^{+}$ becomes noisy (\textit{e.g.}, if we let $\mathbf{k}^{+}$ be target features), queries will be misled into undesirable chaos. 
\item As for queries $\mathbf{q}$, considering the two important characteristics of the introduced pseudo-target domain: (1) similar to the real target domain, and (2) naturally with pixel-wise ground truths (because the translation engine usually keeps the semantic information for each pixel), \textit{sampling $\mathbf{q}$ from the pseudo target domain} is the best choice.
\vspace{-6pt}
\end{itemize}
To this end, this type of alignment works \textit{only} when we pull target to source.
}

\paragraph{The effectiveness of sample strategies} is studied in Tab.~\ref{tab:abla_sampling}.
Notably, only when combining these two strategies together brings significant improvements.
On one hand, without class-balanced query sampling (CBQS), queries are easily overwhelmed by those dominant classes, \textit{e.g.}, ``road'' and ``sky''.
On the other hand, without domain-equalized negative pairs sampling (DENPS), the model tends to neglect target features in contrastive learning.

\paragraph{The effectiveness of dynamic re-weighting} (DRW)
is studied in Tab.~\ref{tab:abla_drw}.
Applying DRW to $\mathcal{L}_{\mathrm{target}}$ rather than $\mathcal{L}_{\mathrm{pull}}$ can be an alternative.
However, as underperforming categories tend to suffer from greater domain shifts, their target pseudo-labels can be noisy.
It is risky to optimize those classes using the cross-entropy loss directly.
To this end, it is better to combine DRW with $\mathcal{L}_{\mathrm{pull}}$, urging the model to put more effort into learning similar cross-domain features for these categories.
From the table, we can tell that \textit{only when DRW is equipped with $\mathcal{L}_{\mathrm{pull}}$}, it brings significant improvements, especially in tailed classes.
%

\begin{figure*}[t]
    \centering
    \includegraphics[width=1\linewidth]{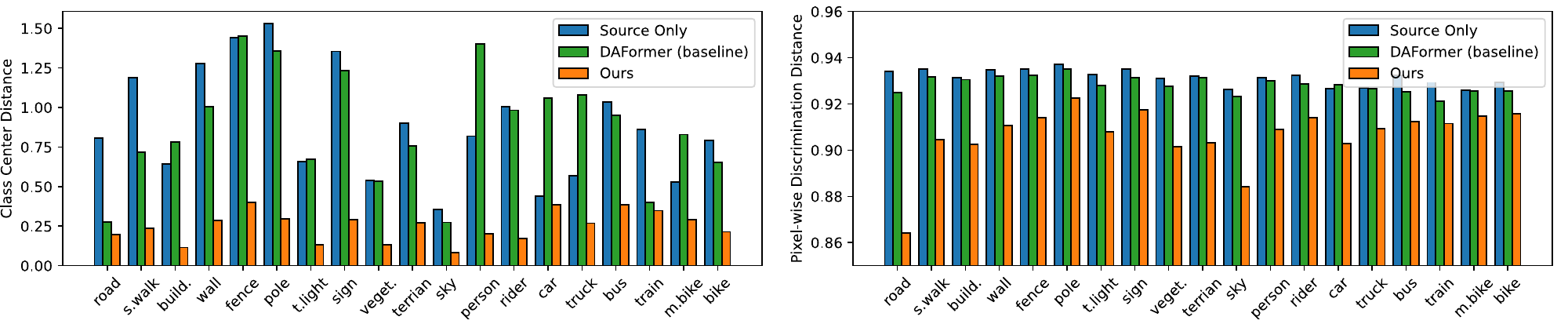}
    \vspace{-10pt}
    \caption{%
        \textbf{Analysis of the feature discriminativeness.}
        For each class, we show the values of class center distance \citep{wang2020classes} and pixel-wise discrimination distance \citep{xie2022sepico} on Cityscapes \texttt{val} set.
        For both metrics, a \textit{smaller} value indicates \textit{stronger} discrimination.
    }
    \label{fig:ccd_pdd}
\end{figure*}

\begin{figure*}[t]
    \centering
    \includegraphics[width=0.9\linewidth]{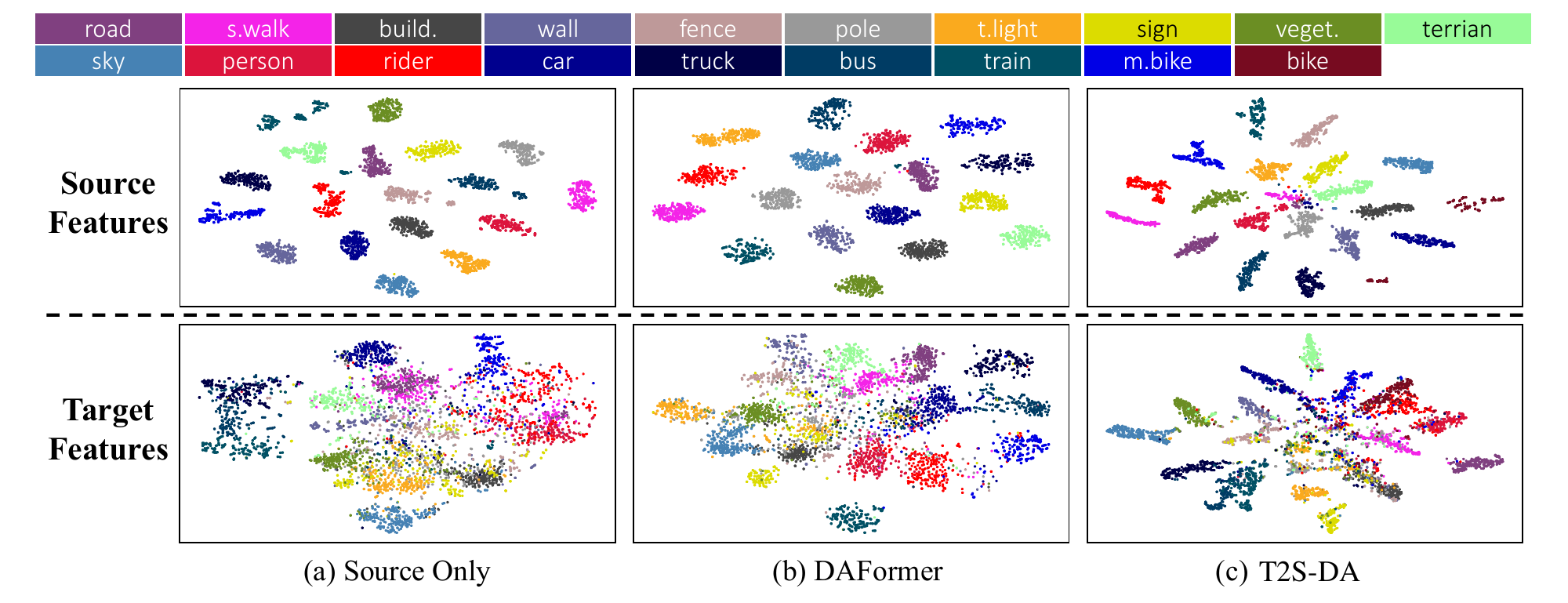}
    \vspace{-8pt}
    \caption{
    Comparison of the feature space using t-SNE \citep{tsne} visualization between (a) the source-only baseline, (b) DAFormer \citep{hoyer2022daformer}, and (c) our proposed \method.
    Note that we randomly sample $256$ pixels for each category for better visualization.
    }
    \label{fig:feature_space}
\end{figure*}

\paragraph{The effectiveness of different objectives} is studied in Tab.~\ref{tab:abla_loss}.
InfoNCE \citep{zou2018unsupervised} is the default alignment objective by default, however, \textit{our motivation is not limited to contrastive learning}.
Interestingly, when replacing the InfoNCE with the MSE loss between $\ell_2$ normalized positive pairs $(\mathbf{q}, \mathbf{k}^+)$, it can still boost the performance with $+2.3\%$ and $+1.1\%$ mIoU improvements using CNN-based models and Transformer-based models, respectively.
Details can be found in the following table.

\paragraph{The effectiveness of the image translation engine} is studied in Tab.~\ref{tab:abla_translation}.
Implementation details of ColorJitter and GaussianBlur are the same with SimCLR \citep{chen2020simple}.
We adopt FDA \citep{yang2020fda} simply because it does not need extra training.
This empirical evidence demonstrates that in our \method, it is not necessary to exactly \textit{fit} the target domain.
In other words, \method learns domain-invariant features \textit{no matter what kind of transformation is adopted}.
\revise{
Moreover, empirical evidence illustrated in Tab.~\ref{tab:abla_translation} indicates that the quality of the translation engine is \textit{not} the bottleneck, but how to leverage it appropriately is, because \textit{little} improvements are observed when applying the source cross-entropy loss on the translated pseudo-target domain.
}

\paragraph{Ablation of $\lambda$.}
Tab. \ref{tab:lambda} shows the performances of our \method with different weights of contrastive loss $\lambda$, among which $\lambda=0.1$ performs the best.
When $\lambda$ is too small (\textit{i.e.}, $0.01$), the performance is close to the baseline, indicating that the target features are still undesirable, while when $\lambda$ is too large (\textit{i.e.}, $1$), the model tends to pay too much attention on learn similar features across domains rather than focus on the particular downstream task: semantic segmentation.

\paragraph{Ablation of $\tau$.}
Tab. \ref{tab:temp} ablates the tunable temperature parameter $\tau$ introduced in Eq. (\textcolor{red}{2}).
We find that $\tau=0.2$ achieves the best performance.
One potential reason is that a smaller $\tau$ provides less difficulty in matching the correct positive key, while a larger $\tau$ results in too strict criteria for this dictionary look-up pretext task, which is similar to having a large contrastive loss weight $\lambda$.

\paragraph{Ablation of $\beta$.}
Tab. \ref{tab:beta} studies the scale parameter of dynamic re-weighting described in Eq. (\textcolor{red}{9}), where $\beta=0.5$ yields slightly better performance, indicating that dynamic re-weighting is quite robust against $\beta$.

\paragraph{Ablation of $\alpha$.}
Tab. \ref{tab:apha} ablates the unreliable partition $\alpha$ introduced in Eq. (\textcolor{red}{7}), which is used to filter unreliable predictions on the target domain to be candidate negative pairs.
As illustrated, $\alpha=50\%$ achieves slightly better performance, and our \method is insensitive to $\alpha$.

\paragraph{Ablation of $n$ and $m$.}
Tab. \ref{tab:n_m} studies the base number of queries per class $n$ and the total number of negative pairs per query $m$ introduced in Sec. \textcolor{red}{3.3}.
We find that $n=128$ and $m=1024$ yield slightly better performance.
Therefore, our framework is found to be stable to different $n$ and $m$.




\begin{table*}[t]
\caption{
Evaluation on the \textbf{source domain} on GTA5 $\to$ Cityscapes benchmark.
\method learns robust domain-invariant features with good generalization abilities since it improves the performance on \textit{both} domains.
}
\setlength{\tabcolsep}{3.2pt}
\label{tab:source}
\vspace{-5pt}
\centering\footnotesize
\begin{tabular}{l | ccccccccccccccccccc | c}
\toprule
Method &
\rotatebox{90}{Road} & \rotatebox{90}{S.walk} & \rotatebox{90}{Build.} & \rotatebox{90}{Wall} & \rotatebox{90}{Fence} & \rotatebox{90}{Pole} & \rotatebox{90}{T.light} & \rotatebox{90}{Sign} & \rotatebox{90}{Veget.} & \rotatebox{90}{Terrain} & \rotatebox{90}{Sky} & \rotatebox{90}{Person} & \rotatebox{90}{Rider} & \rotatebox{90}{Car} & \rotatebox{90}{Truck} & \rotatebox{90}{Bus} & \rotatebox{90}{Train} & \rotatebox{90}{M.bike} & \rotatebox{90}{Bike} & mIoU \\
\midrule
source only & 
97.6 & 89.7 & 91.8 & 72.4 & 58.9 & 66.2 & 66.6 & 68.1 & 86.1 & 76.1 & 96.6 & 79.8 & 70.0 & 92.7 & 90.8 & 94.4 & 87.5 & 73.7 & 53.9 & 79.6  \\
DAFormer & 
97.2 & 87.9 & 91.2 & 68.9 & 55.9 & 65.7 & 67.5 & 71.5 & 85.9 & 74.7 & 96.6 & 79.9 & 75.6 & 92.4 & 89.9 & 93.9 & 92.7 & 80.3 & 64.9 & 80.7 \\
\method (ours) & 
97.2 & 88.1 & 91.3 & 69.2 & 57.0 & 66.6 & 68.2 & 71.4 & 86.0 & 74.8 & 96.6 & 79.7 & 76.0 & 92.4 & 90.2 & 94.7 & 91.2 & 80.4 & 64.7 & 80.9 \\
\bottomrule
\end{tabular}
\end{table*}

\section{Analysis}

\subsection{Feature Discriminativeness Analysis}
To verify whether our \method has indeed built a more discriminative target representation space compared with previous alternatives, we adopt metrics used in FADA \citep{wang2020classes} and SePiCo \citep{xie2022sepico} to take a closer look at what degree the representations are aligned at category-level and pixel-level, respectively.
We compare the discrimination of target feature space between (1) the source-only baseline, (2) DAFormer \citep{hoyer2022daformer}, and (3) our \method.
We calculate these metrics on the whole Cityscapes \citep{cordts2016cityscapes} validation set.
Detailed formulations of these two metrics are given below.

\paragraph{Class center distance (CCD)} is the ratio of intra-class compactness over inter-class distance \citep{wang2020classes}
\begin{equation}
    CDD(i) = \frac{1}{C-1} \sum_{j=0,j\neq i}^{C-1} 
    \frac{
        \frac{1}{|\Omega_i|} \sum_{\mathbf{x} \in \Omega_i} ||\mathbf{x} - \bm{\mu}_i||^2
    }
    {
        ||\bm{\mu}_i - \bm{\mu}_j||^2
    },
\end{equation}
where $\bm{\mu}_i$ is the prototype or center of category $i$ and $\Omega_i$ is the set of features that belong to class $i$.

\paragraph{Pixel-wise discrimination distance (PDD)} is to evaluate the effectiveness of pixel-wise representation alignment \citep{xie2022sepico}.
We modify it into $1$ minus its original value for a more straightforward comparison together with CCD.
\begin{equation}
    PDD(i) = 1 - \frac{1}{|\Omega_i|} \sum_{\mathbf{x} \in \Omega_i} \frac{
        \cos(\mathbf{x}, \bm{\mu}_i)
    }
    {
        \sum_{j=0,j\neq i}^{C-1} \cos(\mathbf{x}, \bm{\mu}_j)
    }.
\end{equation}

Low CCD and PDD suggest the features of the same class have been densely clustered while the distances between different categories are relatively large at category level and pixel level, respectively.
We compare CCD and PDD values with source only model and DAFormer \citep{hoyer2022daformer}.
As illustrated in Fig. \ref{fig:ccd_pdd}, \method achieves much smaller CCD and PDD values in each category compared with others.
This empirical evidence verifies that by pulling target features close to source ones for each category, \method indeed learns a more discriminative target feature space, resulting in better segmentation performances.

\paragraph{Visualization.}
Fig. \ref{fig:feature_space} gives a visualization of feature spaces of (1) the source-only baseline, (2) DAFormer \citep{hoyer2022daformer}, and (3) our proposed \method,
where their source features are discriminative enough but target features for source only baseline and DAFormer \citep{hoyer2022daformer} run into chaos, while our proposed \method is able to build a category-discriminative feature space.
This indicates that only by self-training as DAFormer \citep{hoyer2022daformer} does, target features are not capable enough.
By urging the model to learn similar cross-domain features, our proposed \method makes decision boundaries lie in low-density regions.

\subsection{Performance on Source Domain}

As our proposed \method tries to make target features as similar to source features as possible, leading to a category-discriminative feature representation space on the target domain, 
which implicitly assumes the model will \textit{not} deteriorate and is able to maintain a capable feature space on the source domain during training.
Therefore, we conduct experiments in Tab. \ref{tab:source} to verify this assumption.
We do not find a significant gap between the three methods, inducing the source-only baseline, DAFormer \citep{hoyer2022daformer}, and our proposed \method.

\subsection{Qualitative Results}

\begin{figure*}[t]
    \centering
    \includegraphics[width=1\linewidth]{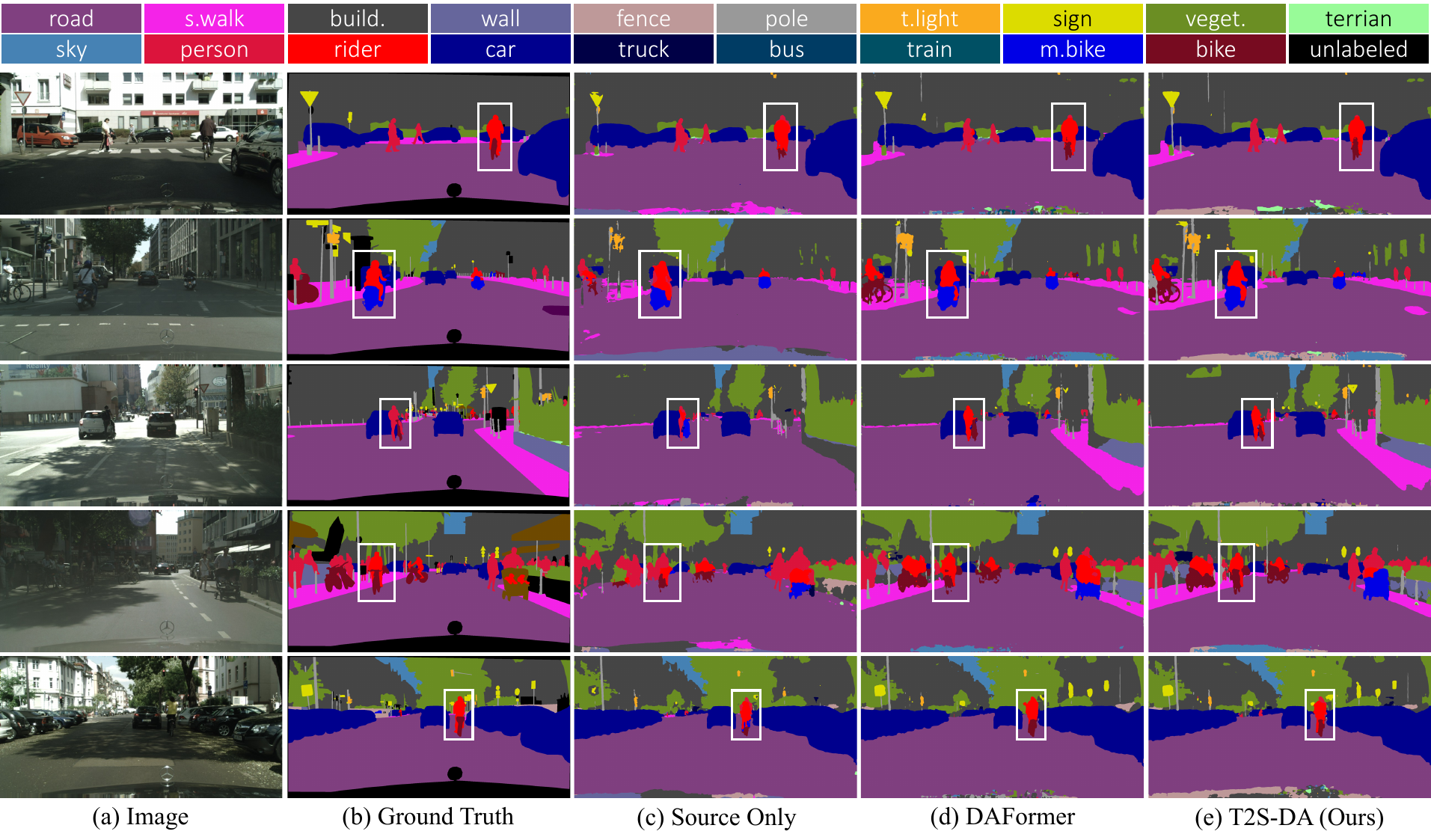}
    \caption{
    Example predictions showing better recognition and finer segmentation details of borders of ``rider'' and ``bike/motorbike'' on GTA5 $\to$ Cityscapes (best-viewed zoomed-in).
    %
    }
    \label{fig:visual_rider}
\end{figure*}

\begin{figure*}[t]
    \centering
    \includegraphics[width=1\linewidth]{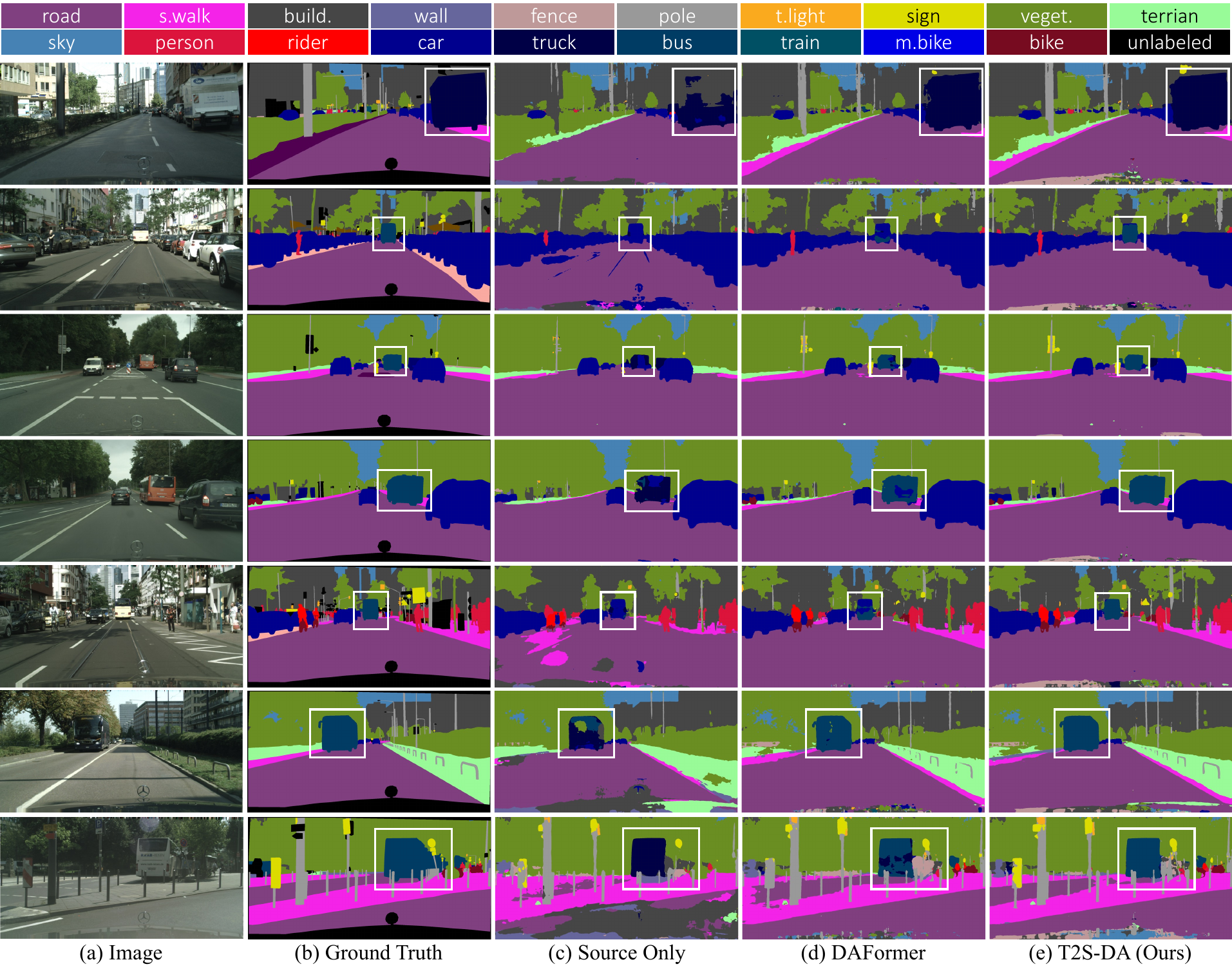}
    \caption{
    Example predictions showing a better recognition and finer segmentation details of ``bus'' and ``truck'' on GTA5 $\to$ Cityscapes,
    which are easily confused with a dominant category ``car'' (best-viewed zoomed-in).
    %
    }
    \label{fig:visual_bus}
\end{figure*}

\begin{figure*}[t]
    \centering
    \includegraphics[width=1\linewidth]{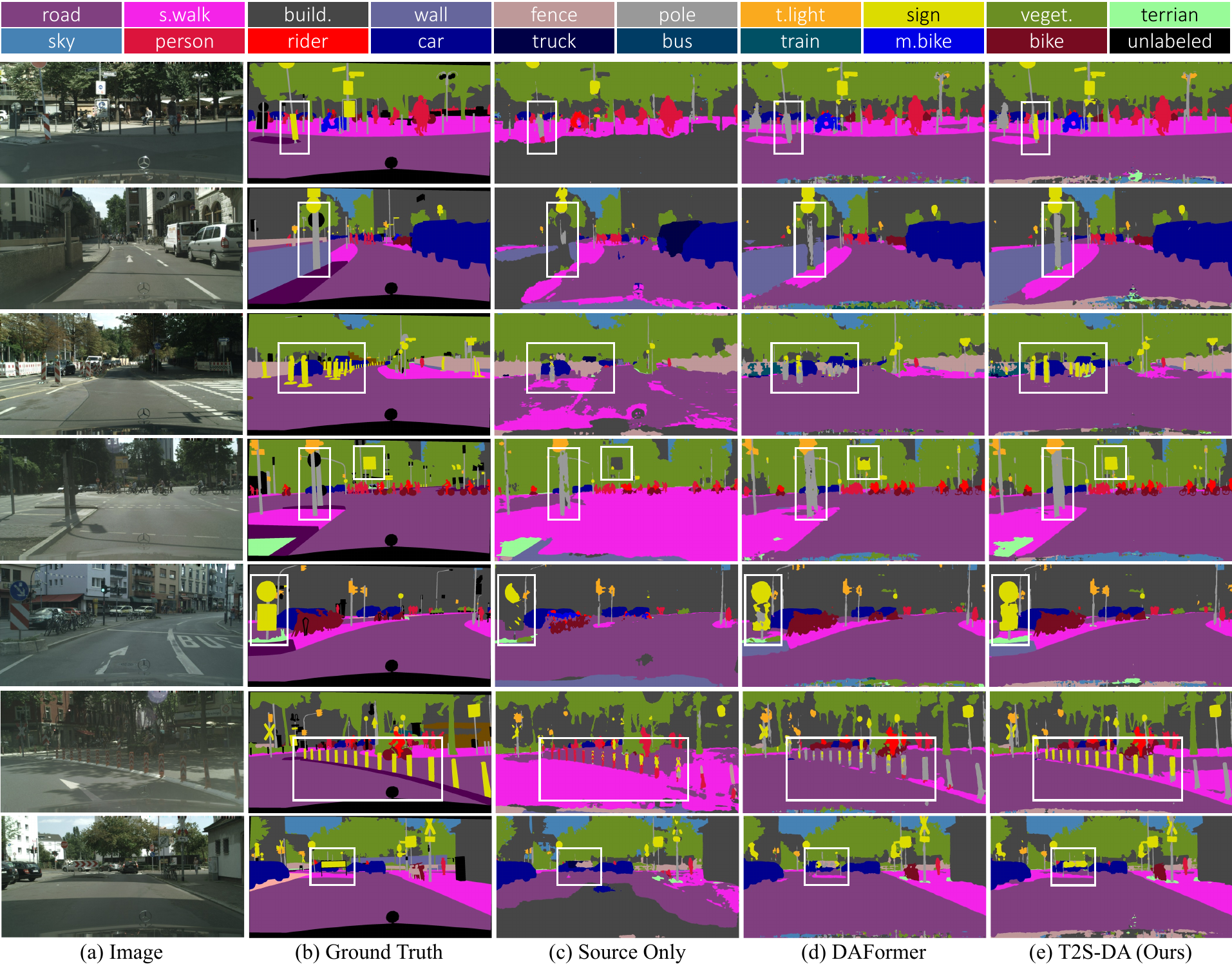}
    \caption{
    Example predictions showing better recognition and finer segmentation details of underperforming categories (``pole'' and ``sign'') on GTA5 $\to$ Cityscapes (best-viewed zoomed-in).
    %
    %
    }
    \label{fig:visual_sign}
\end{figure*}

Fig. \ref{fig:visual_rider}, Fig. \ref{fig:visual_bus}, and Fig. \ref{fig:visual_sign} illustrate visualization results of different methods evaluated on GTA5 $\to$ Cityscapes benchmark.
We compare the proposed \method with ground truths and a strong baseline DAFormer \citep{hoyer2022daformer}.

From Fig. \ref{fig:visual_rider}, we can tell that \method is able to distinguish the conjunction of ``rider'' and ``bike/motorbike'', benefiting from pushing features for different categories away.
From Fig. \ref{fig:visual_bus}, we can tell that \method can bootstrap the performance on categories with large cross-domain feature dissimilarities (\textit{i.e.}, ``bus'' in Fig. \textcolor{red}{1}),
indicating that \method is able to extract domain-invariant feature representations which leads to better performance.
From Fig. \ref{fig:visual_sign}, we can tell that \method pays more attention to underperforming categories, \textit{i.e.}, ``pole'' and ``sign'' where the IoU scores of source only baseline are $34.5\%$ and $23.4\%$ respectively (see Tab. \textcolor{red}{1}), thanks to dynamic re-weighting.

\section{Conclusion}\label{sec:conclusion}

In this paper, we present \method, 
which is able to build a category-discriminative target feature space via \textit{pulling target close to source}, leading to better segmentation results.
%
%
Experimental results show significant growths, and \method proves to be very efficient at avoiding models being trapped in domain-specific styles, outperforming state-of-the-art alternatives on various UDA benchmarks.
Moreover, it can be applied to DG thanks to its domain-invariant property.

\paragraph{Discussion.} 
As discussed in \citep{yang2020fda}, image translation models are unstable, although it is better than using target pseudo-labels directly.
Therefore, how to conduct positive pairs to alleviate the false positive issue without introducing extra noise could be further studied.
Additionally, as the entire real world is the target domain in domain generalization, how to conduct a pseudo-target that is able to \textit{almost} cover the target remains an open problem.
\revise{
Moreover, developing a better style transform model for UDA could be a future work, but it is quite challenging because the engine should be able to (1) generate images similar to the real target domain, and (2) guarantee semantic alignment at the pixel level.
Leveraging a pre-trained fundamental generation model, \textit{e.g.}, Stable Diffusion~\citep{rombach2021highresolution}, while adding segmentation maps as conditions following~\citep{zhang2023adding} may be a solution.
However, making generative models understand semantic information and follow its guidance is still a big challenge.}

\section*{Declarations}

\noindent\textbf{Acknowledgements}
We would like to express our sincere gratitude to all reviewers and the Associate Editor for their valuable comments and suggestions. 
Their insightful feedback helped us to improve the quality of this paper significantly.
\vspace{6pt}

\noindent\textbf{Funding}
This work was supported in part by the National Key R\&D Program of China (No. 2022ZD0116500), the National Natural Science Foundation of China (No. U21B2042), and in part by the 2035 Innovation Program of CAS, and the InnoHK program.
\vspace{6pt}

\noindent\textbf{Data Availability}
The datasets generated during and/or analyzed during the current study are available from the GTA5\footnote{\url{https://arxiv.org/pdf/1608.02192v1.pdf}}, the Synthia\footnote{\url{https://synthia-dataset.net}}, and the Cityscapes\footnote{\url{https://www.cityscapes-dataset.com}}.

%
%


{\footnotesize
    \bibliographystyle{apalike}
    \bibliography{ref.bib}
}

\end{document}